\documentclass{article}
\pdfoutput=1

\usepackage{microtype}
\usepackage{graphicx}
\usepackage{subcaption}
\usepackage{booktabs} 

\usepackage[ruled,algo2e]{algorithm2e}
\usepackage[dvipsnames]{xcolor}
\usepackage{float}
\usepackage{amssymb}
\usepackage{amsthm}
\usepackage{amsmath}
\usepackage{scalerel}
\usepackage{stackengine}
\usepackage[nointegrals]{wasysym}
\usepackage{tablefootnote}
\usepackage{wrapfig}
\usepackage{xfrac}
\usepackage{comment}

\usepackage[nonatbib, final]{neurips_2022}
\usepackage[numbers]{natbib}

\usepackage{thmtools} 
\usepackage{thm-restate}

\usepackage{siunitx}

\usepackage{hyperref}
\hypersetup{
    colorlinks=true,
    linkcolor=magenta,
    citecolor=NavyBlue,
    filecolor=magenta,      
    urlcolor=blue,
    pdfpagemode=FullScreen,
    }

\makeatletter
\newcommand{\removeParBefore}{\ifvmode\vspace*{-\baselineskip}\setlength{\parskip}{0ex}\fi}
\newcommand{\removeParAfter}{\@ifnextchar\par\@gobble\relax}
\newcommand{\eq}{\begingroup\removeParBefore\endlinechar=32 \eqinner}
\newcommand{\eqinner}[2][aligned]{\endlinechar=32%
\begin{gather}\begin{#1}#2\end{#1}\end{gather}\endgroup\removeParAfter}
\makeatother
\usepackage{tabularx}
\usepackage{multirow}
\usepackage{mathtools}
\usepackage{bbm}

\newcommand{\p}{\mathbf{p}}

\usepackage{anyfontsize}

\usepackage{hyperref}

\usepackage{algorithmic}
\usepackage{algorithm}

\newtheorem{theorem}{Theorem}
\newtheorem{lemma}[theorem]{Lemma}

\usepackage[final]{neurips_2022}

\usepackage{cleveref}

\DeclareMathOperator*{\argmax}{arg\,max}

\title{Learning General World Models in a Handful of Reward-Free Deployments}

\author{%
    Yingchen Xu\thanks{Equal contribution. Correspondence to \url{ycxu@meta.com}.} \\ UCL, Meta AI 
    \And Jack Parker-Holder\footnotemark[1] \\ University of Oxford
    \And Aldo Pacchiano\footnotemark[1] \\ Microsoft Research
    \And Philip J. Ball\footnotemark[1] \\ University of Oxford
    \AND Oleh Rybkin  \\ UPenn
    \And Stephen J. Roberts \\ University of Oxford
    \And Tim Rockt\"{a}schel \\ UCL
    \And Edward Grefenstette \\ UCL, Cohere
}

\begin{document}

\maketitle

\begin{abstract}
Building generally capable agents is a grand challenge for deep reinforcement learning (RL). To approach this challenge practically, we outline two key desiderata: 1) to facilitate generalization, exploration should be task agnostic; 2) to facilitate scalability, exploration policies should collect large quantities of data without costly centralized retraining. Combining these two properties, we introduce the \emph{reward-free deployment efficiency} setting, a new paradigm for RL research.
We then present \textsc{cascade}, a novel approach for self-supervised exploration in this new setting. \textsc{cascade} seeks to learn a world model by collecting data with a population of agents, using an information theoretic objective inspired by Bayesian Active Learning. \textsc{cascade} achieves this by specifically maximizing the \emph{diversity of trajectories} sampled by the population through a novel \emph{cascading} objective. We provide theoretical intuition for \textsc{cascade} which we show in a tabular setting improves upon na\"ive approaches that do not account for population diversity. We then demonstrate that \textsc{cascade} collects diverse task-agnostic datasets and learns agents that generalize zero-shot to novel, unseen downstream tasks on Atari, MiniGrid, Crafter and the DM Control Suite. Code and videos are available at  \url{https://ycxuyingchen.github.io/cascade/}
\end{abstract}

\section{Introduction}

Reinforcement learning (RL, \citep{Sutton1998}) has achieved a number of impressive feats over the past decade, with successes in games \citep{dqn,dota,alphago}, robotics \citep{kalashnikov2018scalable,dexterous_openai}, and the emergence of real world applications \citep{loon,rl_plasma}. Indeed, now that RL has successfully mastered a host of individual tasks, the community has begun to focus on the grand challenge of building \emph{generally capable} agents \citep{gato, xland,lexa2021, saycan2022arxiv}. 

In this work, we take steps towards building \emph{generalist agents at scale}, where we outline two key desiderata. First, for agents to become generalists that can adapt to novel tasks, we eschew the notion of restricting agent learning to task-specific reward functions and focus on the \emph{reward-free} problem setting instead \cite{icm, eysenbach2018diversity}, whereby agents must discover novel skills and behaviors without supervision.\footnote{Indeed, designing reward functions to learn behaviors can have unintended consequences \cite{Ng99policyinvariance,amodei2016faulty}.} Consider the problem of learning to control robotic arms, where we may already have some expert offline data to learn from. In many cases this data will cover only a subset of the entire range of possible behaviors. Therefore, to learn additional {general} skills, it is imperative to  collect additional novel and diverse data, and to do so {without} a pre-specified reward function. 

Second, to ensure scalability, we should have access to a large fleet of robots that we can \emph{deploy} to gather this data for a large number of timesteps \citep{armfarm}, without costly and lengthy centralized retraining during this crucial phase \cite{matsushima2021deploymentefficient}. This has recently been referred to as \emph{deployment efficiency}, falling between the typical online/offline RL dichotomy. Limiting deployments not only reduces the overhead in retraining exploration policies, but also limits the potential costs and risks present when deploying new policies \citep{MingCostSens1993,KoberRLRoboticsSurvey2013}, an important consideration in many real world settings, such as robotics \citep{kalashnikov2018scalable}, education \citep{education} and healthcare \citep{NEURIPS2021_26405399}.

Combining these two desiderata, we introduce the \emph{reward-free deployment efficiency} setting, a new paradigm for deep RL research. To tackle this new problem we train a population of exploration policies to collect large quantities of useful data via \emph{world models} \citep{pets,worldmodels,planet}. World models allow agents to plan and/or train policies without interacting with the true environment. They have already been shown to be highly effective for deployment efficiency \citep{matsushima2021deploymentefficient}, offline RL \citep{mopo, morel,argenson2021modelbased, lompo} and self-supervised exploration \citep{shyam2019model,sekar2020planning}. Furthermore, world models offer the potential for increasing agent generalization capabilities \citep{anand2022procedural, hafner2022benchmarking, lexa2021, Byravan2019ImaginedVG, argenson2021modelbased, clavera2018learning, ball2021augwm, lee2020context}, one of the frontiers of RL research \citep{packer2018assessing,kirk2021survey}. However, since existing self-supervised methods for learning world models are designed to collect only a few transitions with a single exploration policy, they likely produce a \emph{homogenous dataset} when deployed at scale, which does not optimally improve the model.

\begin{figure}[t!]
\vspace{-3mm}
  \centering
  \includegraphics[width=0.99\textwidth]{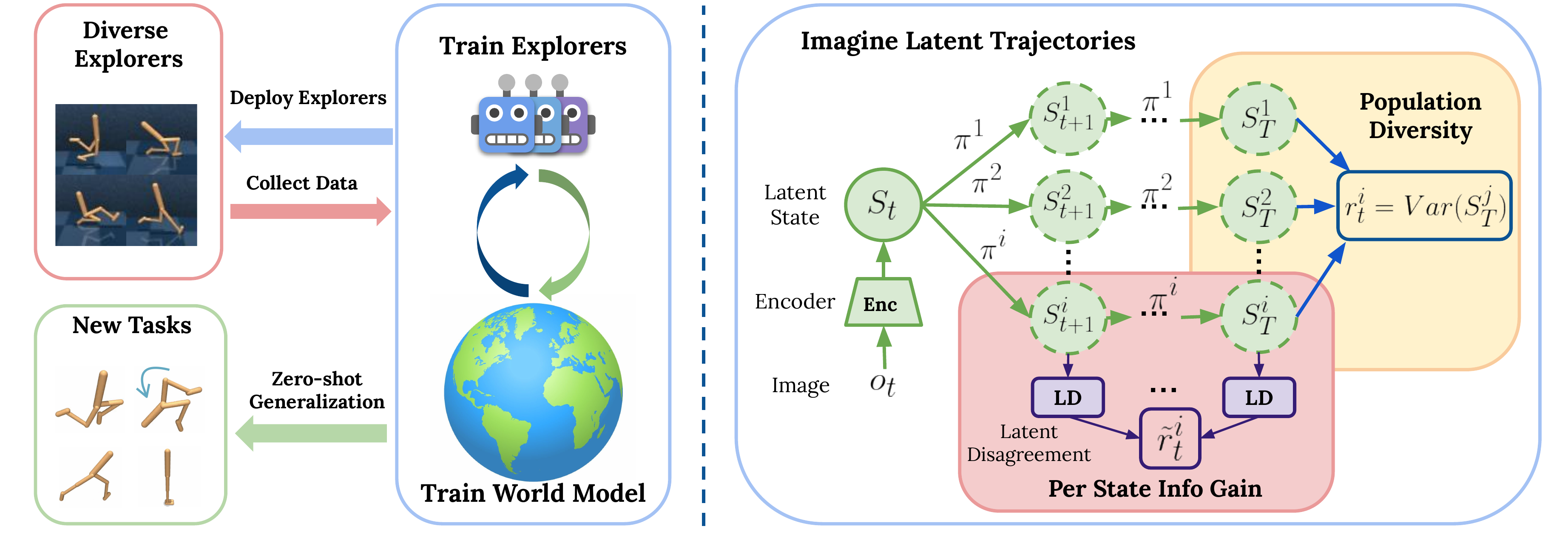}
  \vspace{-2mm}
  \caption{\small{\textbf{Overview.} Left: \textsc{cascade} trains a population of diverse explorers and uses them to collect large batches of reward-free trajectories for learning a general world model that facilitates zero-generalization to novel tasks. Right: To train $B$ exploration agents in parallel, at each training step $t$, \textsc{cascade} first infers a latent state $s_t$ from image observation $o_t$. It then rolls out latent trajectories $\tau^1, \dots, \tau^B$ in imagination using the current exploration policies $\pi^1, \dots,\pi^B$. The training objective for each policy $\pi^i$ is to optimize 1) the population diversity estimated by the disagreement of the final states of imagined trajectories sampled from policies $\pi^1, \dots,\pi^{i}$; 2) the expected per state information gain over all future timesteps $t+1, \dots,T$, computed as the disagreement of an ensemble of dynamics models.}} 
  \vspace{-4mm}
\label{fig:cascade_diagram}  
\end{figure}

\begin{wrapfigure}{r}{0.39\textwidth} 
\vspace{-4.5mm}
    \centering
    \includegraphics[width=0.25\textwidth]{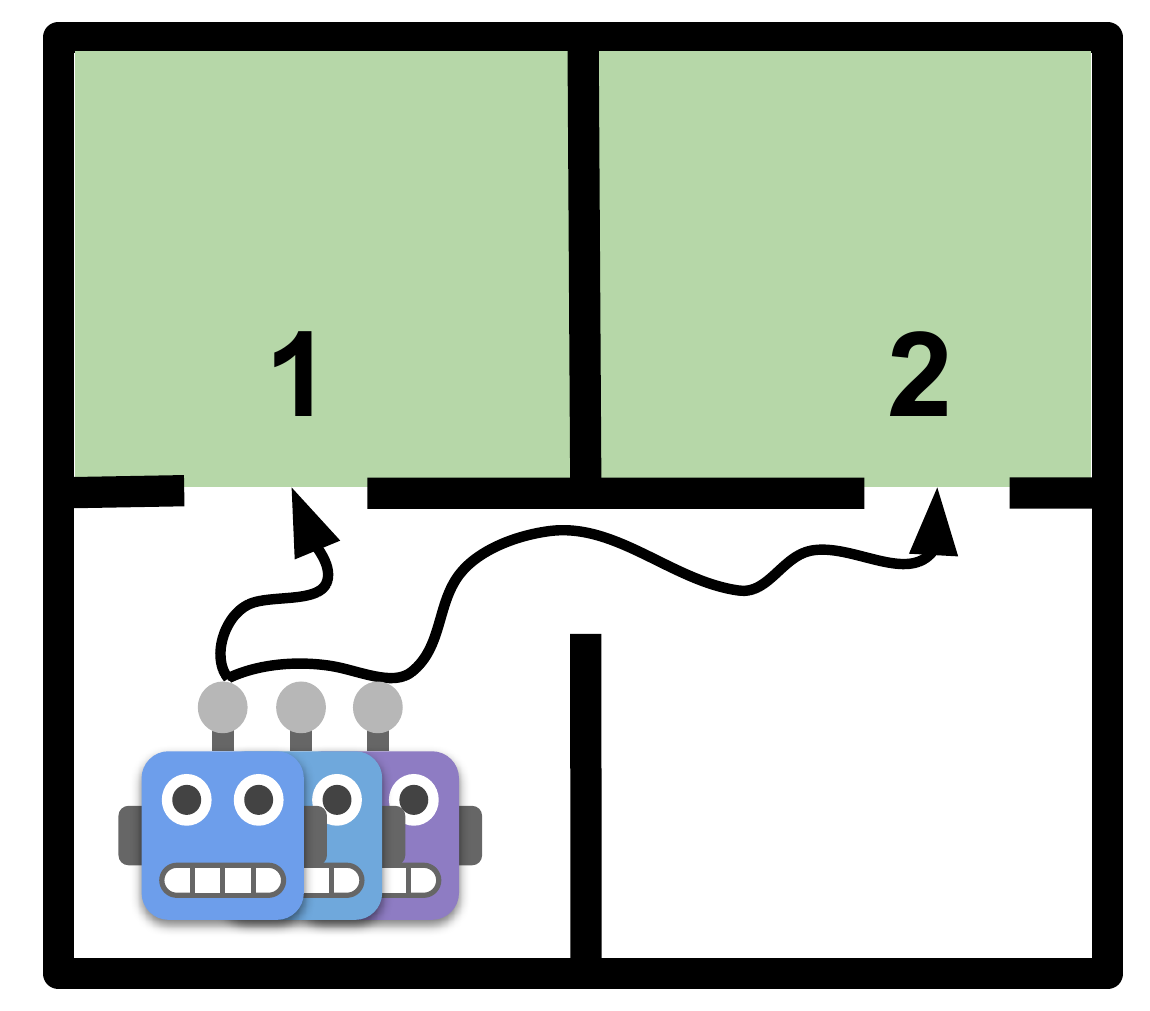}    
    \vspace{-1mm}
    \caption{\small Motivation for \textsc{cascade}: Green areas represent high expected information gain. If we train a population of agents independently, at deployment time they will all follow the trajectory to \#1, producing a homogenous dataset. However, if we consider the diversity of the data then we will produce agents that reach both \#1 and \#2.
    }
    \vspace{-4mm}
    \label{fig:cascade_motivation}
\end{wrapfigure}

Instead, drawing analogies from Bayesian Active Learning \citep{bald, kirsch2019batchbald}, we introduce a new information theoretic objective that maximizes the information gain from an \emph{entire dataset} collected by a \emph{population} of exploration agents (see Figure~\ref{fig:cascade_motivation}). We call our method \emph{\textbf{C}oordinated \textbf{A}ctive \textbf{S}ample \textbf{C}ollection vi\textbf{a} \textbf{D}iverse \textbf{E}xplorers} or \textsc{cascade} (Figure ~\ref{fig:cascade_diagram}). We provide theoretical justification for \textsc{cascade}, which emphasizes the importance of collecting data with diverse agents. In addition, we provide a rigorous empirical evaluation across four challenging domains that shows \textsc{cascade} can discover a rich dataset from a handful of deployments. We see that \textsc{cascade} produces general exploration strategies that are equally adept at both ``deep" exploration problems and diverse behavior discovery. This makes it possible to train agents capable of zero-shot transfer when rewards are provided at test time in a variety of different settings.

To summarize, our contributions are as follows: 1) We introduce a novel problem setting, \emph{Reward-Free Deployment Efficiency}, designed to train generalist agents in a scalable fashion; 2) We propose \textsc{cascade}, a theoretically motivated model-based RL agent designed to gather diverse, highly informative data, inspired by Bayesian Active Learning; 3) We provide analysis that shows \textsc{cascade} theoretically improves sample efficiency over other na\"ive methods that do not ensure sample diversity, and demonstrate that \textsc{cascade} is capable of \emph{improved zero-shot transfer} in four distinct settings, ranging from procedurally generated worlds to continuous control from pixels.

\section{Problem Statement}

Reinforcement learning (RL) considers training an agent to solve a Markov Decision Process (MDP), represented as a tuple $\mathcal{M}=\{\mathcal{S},\mathcal{A},P, R,\rho,\gamma\}$, where $s\in\mathcal{S}$ and $a\in\mathcal{A}$ are the set of states and actions respectively, $P(s' |s, a)$ is a probability distribution over next states given a previous state and action, $R(s, a, s')\rightarrow r$ is a reward function mapping a transition to a scalar reward, $\rho$ is an initial state distribution and $\gamma$ is a discount factor. A policy $\pi$ acting in the environment produces a trajectory $\tau = \{s_1, a_1, \dots, s_H, a_H \}$ for an episode with horizon $H$. Since actions in the trajectory are sampled from a policy, we can then define the RL problem as finding a policy $\pi$ that maximizes expected returns in the environment, i.e. $
    \pi^\star = \arg \max_{\pi} \mathbb{E}_{\tau\sim\pi} [R(\tau)]$.

We seek to learn policies that can transfer to \emph{any} MDP within a family of MDPs. This can be formalized as a \emph{Contextual} MDP \citep{kirk2021survey}, where observations, dynamics and rewards can vary given a context. In this paper we consider settings where only the reward varies, thus, if the test-time context is unknown at training time we must collect data that \emph{sufficiently covers the space of possible reward functions}. Finally, to facilitate scalability, we operate in the deployment efficient paradigm \cite{matsushima2021deploymentefficient}, whereby policy learning and exploration are completely separate, and during a given \emph{deployment}, we gather a large quantity of data without further policy retraining (c.f. online approaches like DER \cite{hasselt2019der}, which take multiple gradient steps \emph{per} exploration timestep in the real environment). Taken together, we consider the \emph{reward-free deployment efficiency} problem. This differs from previous work as follows: 1) unlike previous deployment efficiency work, our exploration is task agnostic; 2) unlike previous reward-free RL work, we cannot update our exploration policy $\pi_{\mathrm{EXP}}$ during deployment. Thus, the focus of our work is on how to train $\pi_{\mathrm{EXP}}$ offline such that it gathers heterogeneous and informative data which facilitate zero-shot transfer to unknown tasks. 

In this paper we make use of model-based RL (MBRL), where the goal is to learn a model of the environment (or \emph{world model} \citep{Schmidhuber90makingthe}) and then use it to subsequently train policies to solve downstream tasks. To do this, the world model needs to approximate both $P$ and $R$. Typically, the model will be a neural network, parameterized by $\psi$, hence we denote the approximate dynamics and reward functions as $P_\psi$ and $R_\psi$, which produces a new ``imaginary'' MDP, $\mathcal{M}_\psi = (\mathcal{S}, \mathcal{A}, P_\psi, R_\psi, \rho)$. We focus on Dyna-style MBRL \cite{dyna}, whereby we train a policy ($\pi_\theta$  parameterized by $\theta$) with model-free RL solely using ``imagined'' transitions inside $\mathcal{M}_\psi$. Furthermore, we can train the policy on a single GPU with parallelized rollouts since the simulator is a neural network \citep{METRPO}. The general form of all methods in this paper is shown in Algorithm~\ref{algorithm:batchexp}, with the key difference being step 5: We aim to update $\pi_{\mathrm{EXP}}$ in the new imaginary MDP $\mathcal{M}_\psi$ such that it continues to collect a large, diverse quantity of reward-free data. Note that $\pi_{\mathrm{EXP}}$ need not be a single policy, but could also refer to a collection of policies that we can deploy (either in parallel or in series), such that $\pi \in \pi_{\mathrm{EXP}}$.

\vspace{-2mm}
\begin{algorithm}[H]
\SetAlgoLined
\caption{Reward-Free Deployment Efficiency via World Models}
\begin{algorithmic}[1]
\STATE \textbf{Input:}  Initial exploration policy $\pi_{\mathrm{EXP}}$ 
\FOR{each deployment} 
  \STATE Deploy $\pi_{\mathrm{EXP}}$ to collect a \emph{large quantity} of \emph{reward-free} data.
  \STATE Train world model on all existing data.
  \STATE Update $\pi_{\mathrm{EXP}}$ in new imaginary MDP $\mathcal{M}_\psi$.
\ENDFOR

\end{algorithmic}
\label{algorithm:batchexp}
\end{algorithm}
\vspace{-3mm}

We focus on learning world models from high dimensional sensory inputs such as pixels \citep{worldmodels, oh2015, simple}, where at each timestep we are given access to an observation $o_t$ rather than a state $s_t$. A series of recent works have shown tremendous success by mapping the observation to  a compact latent state $z_t$ \citep{planet, dreamer, hafner2021mastering}. In this paper we will make use of the model from DreamerV2 \citep{hafner2021mastering}, which has been shown to produce highly effective policies in a variety of high dimensional environments. The primary component of DreamerV2 is a Recurrent State Space Model (RSSM) that 
uses a learned latent state to predict the image reconstruction, reward $r_t$ and discount factor $\gamma_t$. 
Aside from the reward head, all components of the model are trained jointly, in similar fashion to variational encoders (VAEs, \citep{kingma2013auto,rezende2014stochastic}). 
For zero-shot evaluation, we follow \citep{sekar2020planning} and only train the reward head at test time when provided with labels for our pre-collected data, which is then used to train a behavior policy offine. Thus, it is critical that our dataset is sufficiently diverse to enable learning novel, unseen behaviors.

\section{Coordinated Active Sample Collection}

The aim of this work is to train a population of $B$ exploration policies $\{ \pi^{(i)}_{\mathrm{EXP}} \}_{i=1}^B$ such that they collectively acquire data which maximally improves the accuracy of a \emph{world model}. To achieve this, we take inspiration from the information theoretic approach in Plan2Explore \citep{sekar2020planning}, but crucially focus on maximizing information gain over \emph{entire trajectories} rather than per state-action, and hence drop the conditional dependence on state and action (see App.\ \ref{lemma::information_theoretic_perspective} for why this distinction is important):
\begin{equation}\label{eq::single_policy_info_objective_dw}
    \pi_{\mathrm{EXP}} = \argmax_{\pi} \mathcal{I}\left( d_{\mathcal{M}_\psi}^\pi ; \mathcal{M}_\psi \right) = \mathcal{H}(d_{\mathcal{M}_\psi}^\pi) - \mathcal{H}(d_{\mathcal{M}_\psi}^\pi | \mathcal{M}_\psi)
\end{equation}
where $d_{\mathcal{M}_\psi}^\pi$ is the distribution of states visited by the policy $\pi$ in the imaginary MDP $\mathcal{M}_\psi$. This objective produces $\pi_{\mathrm{EXP}}$, a policy whose visitation distribution has a high entropy when computed over model samples, but has low entropy for each individual MDP model (i.e., high epistemic/reducable uncertainty). We think of each model $\mathcal{M}_\psi$ as sampled from a posterior distribution over models given the data. A good exploration policy has low entropy on individual models but large entropy across models, i.e. it is intent in visiting regions of the space where there is large uncertainty about the model transitions. To make this objective more general, we represent the trajectory data collected by a policy with a ``summary'' embedding space \citep{wass, moskovitz2021efficient}. Let $\Phi : \Gamma \rightarrow \Omega$ be a summary function mapping trajectories into this embedding space. $\mathbb{P}^\Phi_\pi[{\mathcal{M}_\psi}]$ denotes the embedding distribution generated by policy $\pi$ in imaginary MDP ${\mathcal{M}_\psi}$.
We can now write the objective from Eq.~\ref{eq::single_policy_info_objective_dw} as follows:
\begin{equation}\label{eq::single_policy_info_objective_embedding}
    \pi_{\mathrm{EXP}} = \argmax_{\pi} \mathcal{I}\left( \mathbb{P}_\pi^{\Phi}[{\mathcal{M}_\psi}] ; {\mathcal{M}_\psi} \right) = \mathcal{H}(\mathbb{P}_\pi^{\Phi}[{\mathcal{M}_\psi}]) - \mathcal{H}(\mathbb{P}_\pi^{\Phi}[{\mathcal{M}_\psi}] | {\mathcal{M}_\psi})
\end{equation}
This more general framework allows us to consider multiple representations for trajectories, in a similar fashion to behavioral characterizations in Quality Diversity algorithms \citep{qdnature}. For the rest of this discussion, we will use the final state embedding as our summary representation, whereby $\Phi(\tau) = h_H$, since in the case of the RSSM, the final latent is a compact representation of the entire trajectory collected by the policy, analogous to the final hidden state in an RNN \citep{rnn, seq2seq}.

\subsection{A Cascading Objective with Diverse Explorers}

We now consider a population-based version of Equation.~\ref{eq::single_policy_info_objective_embedding}, using $B$ agents:
\resizebox{.99\linewidth}{!}{
  \begin{minipage}{\linewidth}
    \begin{equation}\label{equation::batch_mutual_info_objective}
    \small{
         \{\pi^{(i)}_{\mathrm{EXP}}\}_{i=1}^B = \argmax_{ \boldsymbol{\pi}^B \in \Pi^B} \mathcal{I}\left( \prod_{i=1}^B \mathbb{P}_{\pi^{(i)}}^{\Phi}[{\mathcal{M}_\psi}] ; {\mathcal{M}_\psi} \right) = \mathcal{H}\left(\prod_{i=1}^B \mathbb{P}_{\pi^{(i)}}^{\Phi}[{\mathcal{M}_\psi}] \right) - \mathcal{H}\left(\prod_{i=1}^B \mathbb{P}_{\pi^{(i)}}^{\Phi}[{\mathcal{M}_\psi}]  \Big| {\mathcal{M}_\psi} \right)
         }
    \end{equation}
  \end{minipage}
}
where $\boldsymbol{\pi}^{B} = \pi^{(1)}, \cdots, \pi^{(B)}$ and $\prod_{i=1}^B \mathbb{P}_{\pi^{(i)}}^{\Phi}[{\mathcal{M}_\psi}] $ is the product measure of the policies' embedding distributions in $\mathcal{M}_\psi$. By definition, the conditional entropy factorizes as:
\begin{equation}\label{equation::entropy_decomposition_product}
    \mathcal{H}\left(\prod_{i=1}^B \mathbb{P}_{\pi^{(i)}}^{\Phi}[{\mathcal{M}_\psi}] \Big| {\mathcal{M}_\psi}\right) = \sum_{i=1}^B \mathcal{H}\left(\mathbb{P}_{\pi^{(i)}}^{\Phi}[{\mathcal{M}_\psi}] \Big |  {\mathcal{M}_\psi}\right).
\end{equation}
It is now possible to show that maximum information gain is achieved with a \emph{diverse} set of agents:

\begin{restatable}{lemma}{lemmadiversityone}\label{lemma::all_policies_same_suboptimality}
When all models $\mathcal{M}_\psi$ in the support of the model posterior are deterministic and tabular, and the space of policies $\Pi$ consists only of deterministic policies, there always exists a solution $   \{\pi^{(i)}_{\mathrm{EXP}}\}_{i=1}^B $ satisfying $\pi_{\mathrm{EXP}}^{(i)} \neq \pi_{\mathrm{EXP}}^{(j)} \forall i \neq j$. Moreover, there exists a family of tabular MDP models, such that the maximum cannot be achieved by setting $\pi_{\mathrm{EXP}}{(i)} = \pi$ for a fixed $\pi$. 
\end{restatable}

The proof of Lemma~\ref{lemma::all_policies_same_suboptimality} is in Appendix~\ref{sec:prooflemma1}. Since the mutual information objective~\ref{equation::batch_mutual_info_objective} is submodular, a greedy algorithm yields a $(1-\frac{1}{e})$ approximation of the optimum (where $e$ is Euler's number)~\cite{nemhauser1978analysis}.
Leveraging this insight, let us assume that we already have a set of policies $\pi^{(1)}, \cdots, \pi^{(i-1)}$; we then select the next policy $\pi^{(i)}$ based on the following greedy objective:

\resizebox{.99\linewidth}{!}{
  \begin{minipage}{\linewidth}
    \begin{align*}
        \pi^{(i)} &= \argmax_{\tilde \pi^{(i)} \in \Pi} \mathcal{I}\left( \prod_{j=1}^i \mathbb{P}_{\tilde \pi^{(j)}}^{\Phi}[{\mathcal{M}_\psi}] ; {\mathcal{M}_\psi} \Big|\tilde \pi^{(j)} = \pi^{(j)}~~ \forall j \leq i-1 \right) \label{equation::greedy_information_theoretic_objective_2}\\
        &=\mathcal{H}\left(\prod_{j=1}^i \mathbb{P}_{\tilde \pi^{(j)}}^{\Phi}[{\mathcal{M}_\psi}] \Big| \tilde\pi^{(j)} = \pi^{(j)}~~ \forall j \leq i-1 \right) - \mathcal{H}\left(\prod_{j=1}^i \mathbb{P}_{\pi^{(j)}}^{\Phi}[\mathcal{M}_\psi]  \Big| \mathcal{M}_\psi , \tilde\pi^{(j)} = \pi^{(j)}~~ \forall j \leq i-1\right)
    \end{align*}
  \end{minipage}
}

Which can be factorized in similar fashion to Equation~\ref{equation::entropy_decomposition_product} (See Appendix \ref{app:deriving_obj}).

\subsection{A Tractable Objective for Deep RL}

Inspired by~\cite{sekar2020planning}, we make a couple of approximations to derive a tractable objective for $\{ \pi_{\mathrm{EXP}}\}_{i=1}^B$ in the deep RL setting. First, we assume that the final state embedding distributions are Gaussian with means that depend on the policies and sampled worlds, and variances that depend on the worlds, i.e.  $\mathbb{P}_{\pi}^{\Phi}[{\mathcal{M}_\psi} = w] = \mathcal{N}(\mu(w, \pi), \Sigma(w))$ . In this case, $\mathcal{H}(\mathbb{P}_{\pi}^{\Phi}[{\mathcal{M}_\psi}] | {\mathcal{M}_\psi} = w ) = \rho(w) $, and Eq.~\ref{equation::greedy_information_theoretic_objective_2} reduces to solving $ \pi^{(i)}  = \arg\max_{\tilde \pi^{(i)} \in \Pi} \mathcal{H}\left(\prod_{j=1}^i \mathbb{P}_{\tilde \pi^{(j)}}^{\Phi}[{\mathcal{M}_\psi}] \Big| \tilde\pi^{(j)} = \pi^{(j)}~~ \forall j \leq i-1 \right)$ for a policy that maximizes the resulting joint entropy of the embedding distribution when added to the policy population. This produces the following surrogate objective, maximizing a quadratic \emph{cascading} disagreement:
\begin{equation*}
       \textcolor{orange}{\mathrm{PopDiv}^\Phi(\pi | \pi^{(1)}, \cdots, \pi^{(i-1)})} = \mathbb{E}_{\tau \sim \mathbb{P}^{\pi}[{\mathcal{M}_\psi}]}\left[ \frac{1}{|\mathcal{D}^{(i-1)}|-1} \sum_{\tilde{\tau} \in \mathcal{D}^{(i-1)}} \left\| \Phi(\tau) -\Phi(\tilde{\tau}) \right\|^2\right]
\end{equation*}
where $\mathcal{D}^{(i-1)}$ is a dataset of imagined trajectories sampled from policies $\pi^{(1)}, \cdots, \pi^{(i-1)}$ in the model, and \textcolor{orange}{$\mathrm{PopDiv}$} is short for Population Diversity.
Finally, following \citep{sekar2020planning, rp1}, we also add a per state information gain component to each policy's reward to encourage a richer landscape for data acquisition: $\textcolor{blue}{\mathrm{InfoGain}(\pi)} = \mathbb{E}_{\tau~\sim \mathbb{P}^{\pi}[{\mathcal{M}_\psi}]}\left[ \sum_{(s,a) \in \tau}\sigma(s,a) \right]$ where $\sigma(\cdot, \cdot)$ is the variance across the ensemble latent state predictions (for details see App.\ \ref{app:imlementation_details}).

Taken together, these objectives form our approach, which we call \emph{\textbf{C}oordinated \textbf{A}ctive \textbf{S}ample \textbf{C}ollection vi\textbf{a} \textbf{D}iverse \textbf{E}xplorers} or \textsc{cascade}. \textsc{cascade} trains agents to optimize: 1) a diversity term (\textcolor{orange}{$\mathrm{PopDiv}$}) that takes into account the behaviors of the other agents in the population; 2) an information gain term (\textcolor{blue}{$\mathrm{InfoGain}$}) that encourages an individual agent to sample states that maximally improve the model:
\begin{align}
    \pi^{(i)} = \argmax_{\pi \in \Pi}\left[ \lambda \textcolor{orange}{
    \mathrm{PopDiv^\Phi(\pi|\{\pi^{(j)}_{\mathrm{EXP}}\}_{j=1}^{i-1})}} + (1-\lambda)\textcolor{blue}{\mathrm{InfoGain}(\pi)} \right]
\end{align}
where $\lambda$ is a weighting hyperparameter that trades off whether we favor individual model information gain or population diversity.
Finally, we train the $B$ agents in \emph{parallel} using (policy) gradient descent over $\theta$, which makes it possible to achieve the same wall clock time as training a single agent \citep{cui2021klevel}.

\subsection{Theoretical Motivation}

We now seek to provide a tabular analogue to \textsc{cascade} which provides a theoretical grounding for our approach.
In App.~\ref{app:theo_rl} we outline the pseudo-code of \textsc{cascade-ts}, a greedy Thompson Sampling algorithm \cite{agrawal2012analysis} that produces the $i$-th exploration policy in a tabular enviornment using ``imaginary" data gathered by running policies $\pi^{(1)}, \cdots, \pi^{(i-1)}$ in the 
\emph{model}. We can then show the following:

\begin{lemma}\label{lemma::cascade_TS}
For the class of Binary Tree MDPs, the \textsc{cascade-ts} algorithm satisfies, 
\begin{equation*}
    T(\epsilon, \mathrm{Sequential}) \leq T(\epsilon, \text{\textsc{cascade-ts}}) \leq T(\epsilon,\mathrm{SinglePolicyBatch} )
\end{equation*}
where $ T(\epsilon, \cdot)$ are the expected number of rounds of deploying a population of $B$ policies necessary to learn the true model up to $\epsilon$ accuracy; $\mathrm{SinglePolicyBatch}$ plays a fixed policy $B$ times in each round; $\mathrm{Sequential}$ does not have a population, and instead interleaves updates and executions of a single policy $B$ times within each round.
\end{lemma}

The proof is in App.~\ref{app:theo_rl}. Indeed, we see that \textsc{cascade-ts} achieves provable efficiency gains over a na\"ive sampling approach that does not ensure diversity in its deployed agents. This provable gain is achieved by discouraging the $i$-th policy away from imaginary state-action pairs sampled by the previous $i-1$ policies by using imaginary counts.

Now returning to \textsc{cascade}, we can see the importance of leveraging the imaginary data gathered in the model by the previous $i-1$ policies when training the $i$-th exploration policy. Concretely, encouraging policy $\pi^{(i)}$ to induce high disagreement with the embeddings produced by $\{\pi^{(1)}, \cdots, \pi^{(i-1)}\}$ (i.e. the $\mathrm{PopDiv}$ term) is analogous to the imaginary count bonus term of \textsc{cascade-ts} in Line 10 of Alg.~\ref{alg:diversity_sequential_algorithm}, which avoids redundant data collection during deployment.

\section{Experiments}

In our experiments we test whether \textsc{cascade} facilitates the learning of generalist agents by evaluating their zero-shot transfer to unseen tasks, given a limited number of reward-free deployments. In all experiments, agents cannot train online in the environment, and instead execute a fixed exploration policy to gather new data during each deployment. We test against three baselines. The first is random exploration, which is used in the majority of RL publications. Our next baseline is Plan2Explore \citep{sekar2020planning} (\textsc{p2e}), which trains a single exploration policy that optimizes for expected information gain inside a world model. We then include a population-based version of \textsc{p2e}, which we call \emph{Population Plan2Explore} (\textsc{pp2e}). \textsc{pp2e} trains a set of randomly initialized agents that independently seek to maximize expected information gain. All methods make use of a DreamerV2 world model \citep{hafner2021mastering} and use the same hyperparameters for model and agent training (more details in App.~\ref{app:imlementation_details}). 

We test all methods in two separate settings: Firstly \emph{Exploring Worlds} (Sec~\ref{sec:expl_worlds}), where we seek to collect data in challenging exploration environments. We test how well the explorers cover the state space and discover sparse rewards (also known as ``deep exploration" \cite{osband2019deepexplore, ODonoghue2020Making}). Secondly, we consider \emph{Exploring Behaviors} (Sec~\ref{sec:expl_behavs}) where we consider collecting data in continuous control environments. In both settings we test zero-shot transfer as follows: 1) we provide reward labels to the model; 2) we train a separate reward head; 3) we train a task specific behavior policy and test it in the environment. This tests whether our model is general enough to facilitate learning downstream policies for previously unknown tasks \citep{sekar2020planning,offlinedatalambert,offlinedatayarats}.

\begin{figure}[H]
    \centering
    \vspace{-1mm}
    \includegraphics[width=1.0\textwidth]{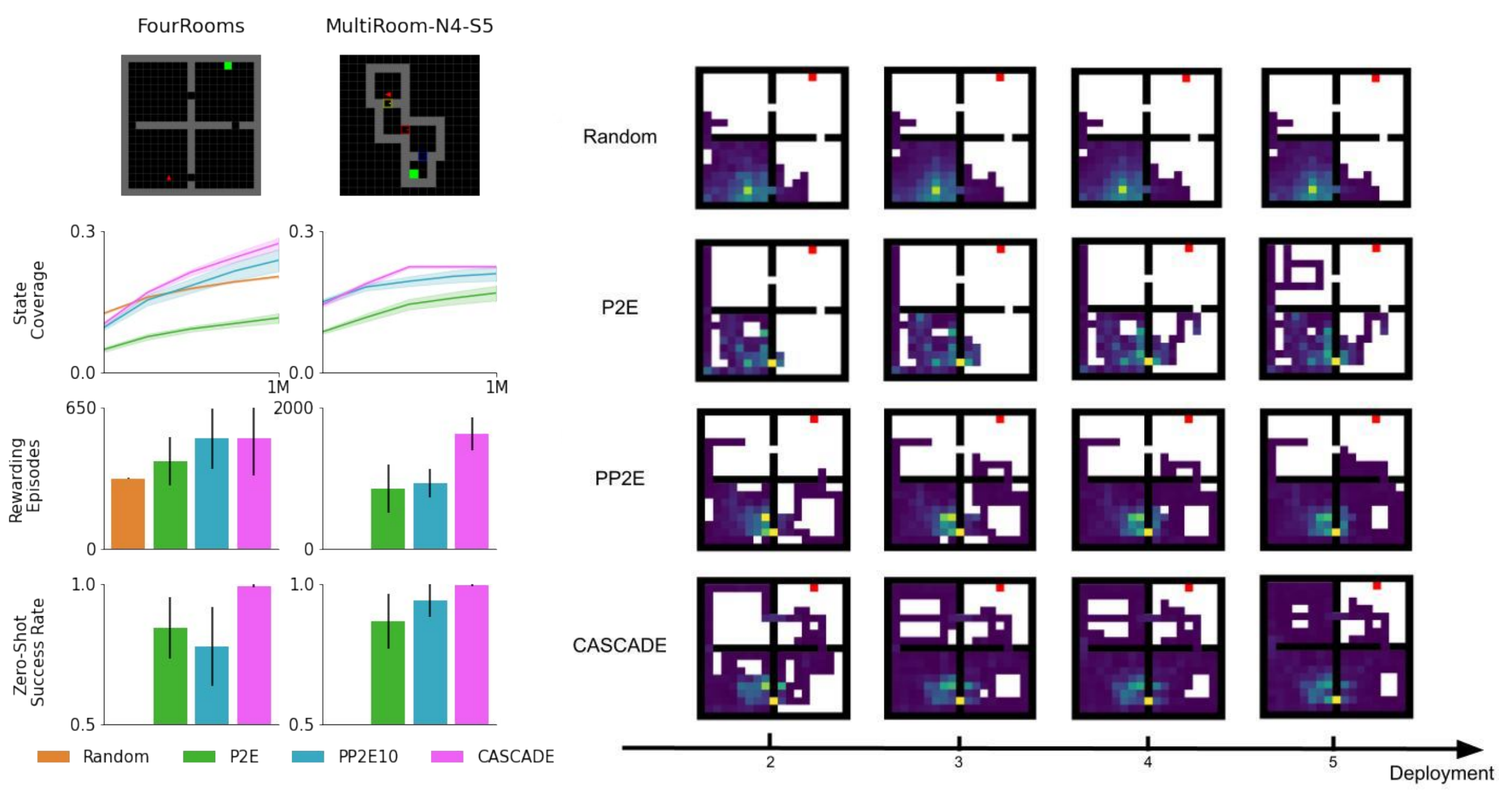}
    \vspace{-5mm}
    \caption{\small{\textbf{MiniGrid Results}. Left: Performance statistics for both \texttt{FourRooms} and \texttt{MultiRoom}. For reward-free exploration we show both ``state coverage'', which corresponds to the the percentage of states visited on a fixed set of levels, and ``rewarding episodes'' which is the count of total solved episodes found after 1M steps. Finally for zero-shot transfer we show the success rate of a task-specific behavior policy trained with labels provided only at test time. All plots show the mean of ten seeds with SEM shaded. (Random achieves 0 zero-shot success rate). Right: we show the state coverage on a fixed level in \texttt{FourRooms}, using the exploration policy $\pi_\mathrm{EXP}$ for each deployment (Test time goal state indicated by the \textcolor{red}{red} dot).}}
    \label{fig:minigrid}
    \vspace{-2mm}
\end{figure}

\subsection{Exploring Worlds}
\label{sec:expl_worlds}

In many cases we may not know a priori which states in an environment are highly rewarding. Thus, if we wish to use our world model to subsequently train agents to solve tasks, it is crucial that we accurately model as much of state space as possible. Here we consider exploring worlds in three different environments: MiniGrid \citep{gym_minigrid}, Atari \citep{ale} and Crafter \citep{hafner2022benchmarking}. In all three settings, we start with a \emph{random} deployment before beginning reward-free exploration.

We begin with MiniGrid, a set of sparse-reward, partially observable navigation environments commonly used to test state-of-the-art exploration methods \citep{ride, amigo, agac, zhang2021noveld}. MiniGrid environments are procedurally generated, which provides an additional generalization challenge \citep{pcg_illuminating, pcg}. We consider the canonical \texttt{FourRooms} and the more challenging \texttt{MultiRoom}. In each case we are limited to 5 deployments of 200k transitions, for a total of 1M environment interactions.\footnote{Zero-shot success rates of \textsc{p2e}, \textsc{pp2e} and \textsc{cascade} increase with more training steps, and \textsc{cascade} achieves a near 100\% success rate at 1M steps in both \texttt{FourRooms} and \texttt{MultiRoom}.} In order to gauge the effectiveness of the reward-free exploration, we use two metrics: 1) state coverage, where we evaluate the percentage of the states visited by the exploration policies after each deployment on the same set of held out test levels; and 2) rewarding episodes, where we show the number of rewarding episodes collected during the training process. Note that these two objectives are not useful in isolation, as solved levels may only include a narrow set of observations, while state coverage alone could be maximized without performing deep, goal-seeking exploration in the environment. However, in combination they provide a proxy for exploration.

We show the results in Figure~\ref{fig:minigrid}, where we see the \textsc{cascade} exploration policies cover the state space far more effectively than all baselines. In \texttt{FourRooms}, both  \textsc{cascade} and  \textsc{pp2e} find the goal an equal number of times, while \textsc{cascade} finds almost double the number of rewarding episodes in the \texttt{MultiRoom} environment. We also test zero-shot performance, where \textsc{cascade} achieves almost 100\% success rate. Interestingly, in both settings the Random policy does cover a large proportion of the state space, but finds fewer rewarding episodes and subsequently does not have a sufficient quantity to train a policy to solve the task (and accordingly achieves 0\% success rate).

\begin{figure}[H]
    \centering
    \includegraphics[width=0.99\textwidth]{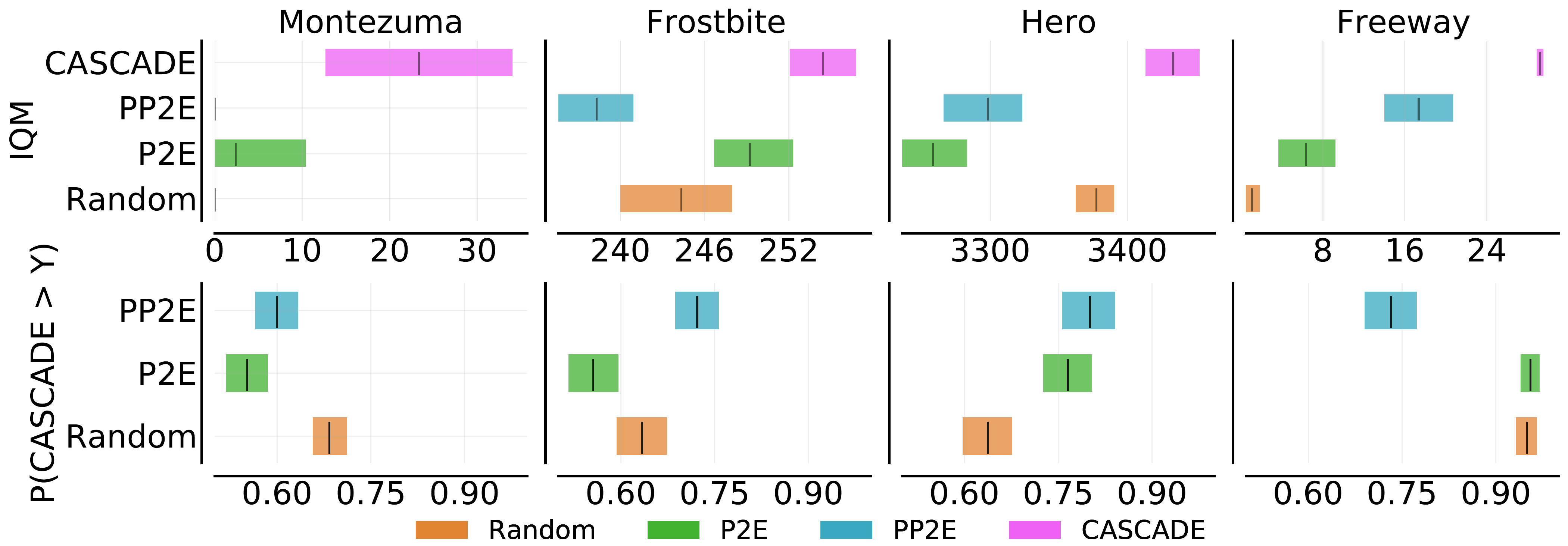}
    \vspace{-1mm}
    \caption{\small{\textbf{Atari Zero-Shot Transfer}: Plots show RLiable metric performance aggregated across all seeds. Note that the `Probability of Improvement' lower CIs all exceed 0.5 for Atari, indicating \textsc{cascade} provides a statistically significant improvement (under the Mann-Whitney U test) over all baseline methods \citep{agarwal2021deep}. We show five seeds of zero-shot test performance.}}
    \vspace{-4mm}
    \label{fig:rliable_atari}
\end{figure}

Next we consider four Atari games: \texttt{Montezuma's Revenge}, \texttt{Frostbite}, \texttt{Hero} and \texttt{Freeway}, often used as barometers for exploration capabilities \citep{Ecoffet2021FirstRT}. We use 15 reward-free deployments, each consisting of 200k steps, and use the final model to train a behavior policy with task reward. We test these agents in a zero-shot fashion, with results shown in the first three columns of Figure~\ref{fig:rliable_atari}. We see that \textsc{cascade} gets statistically significantly stronger zero-shot returns compared to the baselines. Furthermore, the baseline methods display no clear trend, with random exploration sometimes offering the strongest performance. On the other hand, we see that \textsc{cascade} performs consistently well, providing further evidence of it being a strong general exploration strategy. See App.\ \ref{app:atari_deppdive} for more detailed analysis.

\begin{figure}[t!]
    \centering
    \includegraphics[width=0.9\textwidth]{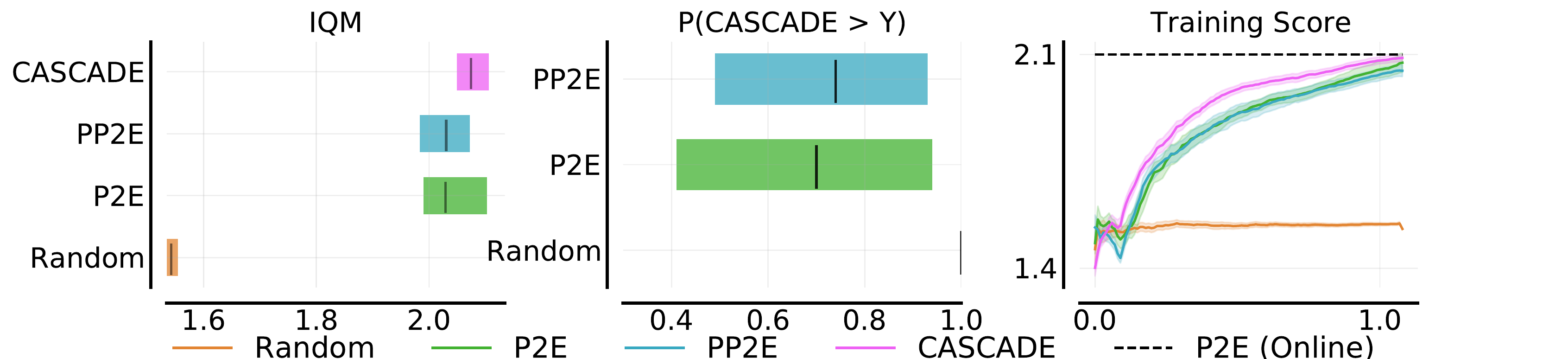}
    \vspace{-1mm}
    \caption{\small{\textbf{Crafter Score}: Plots show final RLiable metric performance and training curve over 1M steps aggregated across all seeds. We show ten seeds of Crafter Score, the geomean of skills discovered in a purely reward-free fashion. Again, see that the `Probability of Improvement' lower CIs all exceed 0.5, indicating \textsc{cascade} provides a statistically significant improvement (under the Mann-Whitney U test) over all baseline methods \citep{agarwal2021deep}.}}
    \vspace{-4mm}
    \label{fig:rliable_crafter}
\end{figure}

\begin{figure}[t!]
    \centering
    \includegraphics[width=0.99\textwidth]{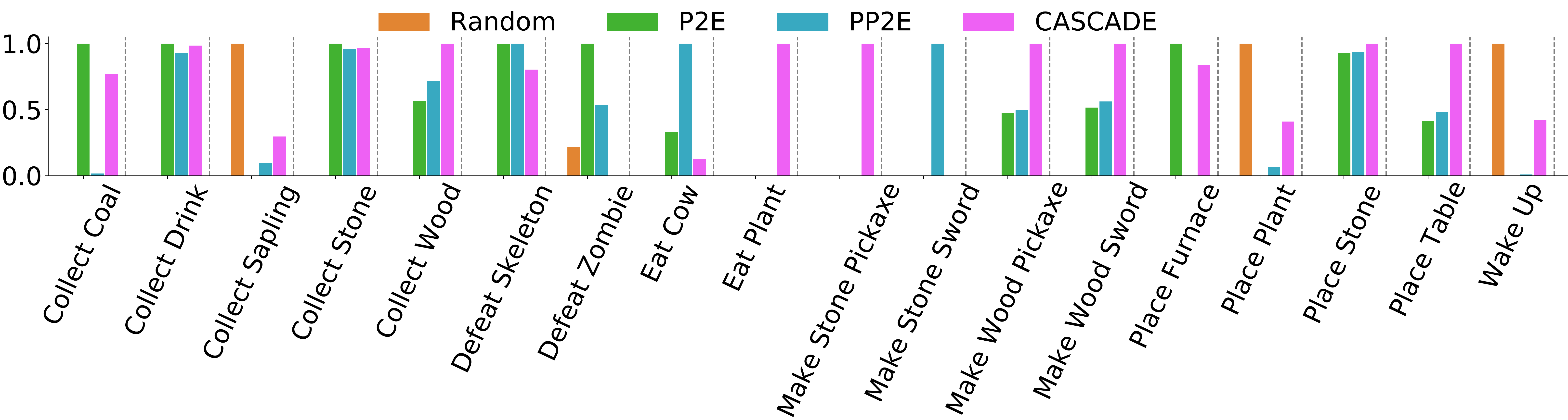}
    \vspace{-2mm}
    \caption{\small{\textbf{Crafter Skills}: Plots show normalized success rate per skill. Note that \textsc{cascade} not only achieves the highest average success rate but also unlocks the most (16 out of 22) unique skills among all baselines. Skills that are not unlocked by any of the baselines are excluded from the plot. }}
    \vspace{-4mm}
    \label{fig:crafter_breakdown}
\end{figure}

Finally, we consider the \texttt{Crafter} environment \citep{hafner2022benchmarking}, a procedurally generated grid world based on the game of Minecraft. This more complex setting requires the agent to master a variety of compositional skills to fully explore all possible behaviors. Online \textsc{p2e} represents the state of the art reward-free agent, achieving a ``Crafter Score'' of 2.1. Making the problem more challenging, we deploy only 20 distinct exploration policies, each collecting 50k timesteps, which leads to weaker performance for \textsc{p2e} vs. the reported result in \cite{hafner2021mastering}. However, as we see in Figure~\ref{fig:rliable_crafter}, adding a population of diverse explorers  recovers the majority of this performance. We find this result encouraging as the behaviors required to achieve a higher Crafter score are non-trivial. Moreover, as shown in Figure~\ref{fig:crafter_breakdown}, \textsc{cascade} achieves the highest average success rate and unlocks the most number of unique Crafter skills among all baselines, including Making a Stone Pickaxe. We believe this opens the door to further gains with more active tuning of the behavioral representation used in \textsc{cascade}.

\subsection{Exploring Behaviors}
\label{sec:expl_behavs}

\begin{figure}[h!]
    \centering
    \vspace{-2mm}
    \includegraphics[width=0.9\textwidth]{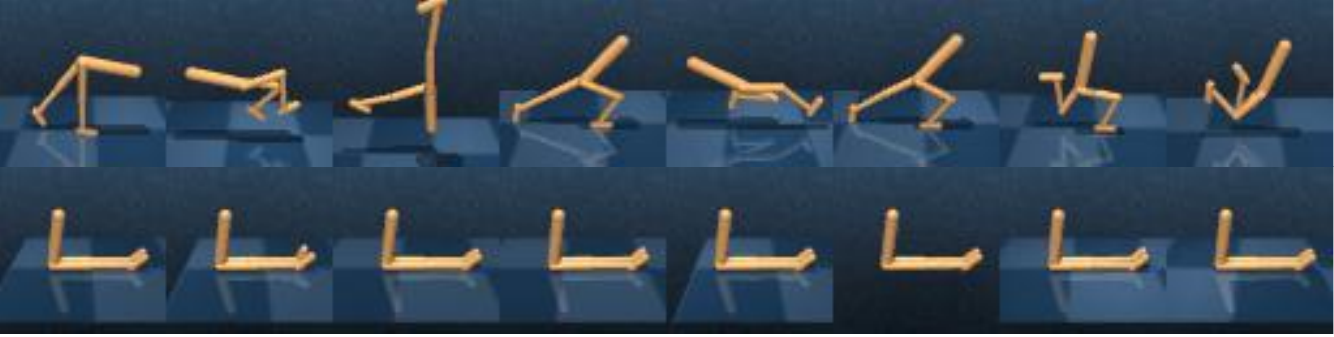}
    \vspace{-1mm}
    \caption{\small{\textbf{DMC Qualitative Results}: Each row visualizes behaviors of a population of explorers trained after 15 total deployments by the same world model, deployed at the same time in environments with the same initial state. We can see that \textsc{cascade} explorers (top) collect data with a diversity of behaviors while \textsc{pp2e} agents (bottom) only show less interesting homogenous behaviors.}}
    \vspace{-2mm}
    \label{fig:walker_qualitative}
\end{figure}

Next we wish to test is whether our agents sufficiently explore a range of behaviors in a continuous control setting, using the \texttt{walker} environment from the DM Control Suite \citep{tunyasuvunakool2020} (DMC). In these experiments, we operate from pixels and assume access to an initial dataset of 200k transitions. Then, we conduct 14 further reward-free deployments, each collecting 200k transitions. We consider the possibility of improving generality from arbitrary offline data, using common datasets from the offline RL literature: \texttt{random}, \texttt{medium} and \texttt{expert}. Using random data represents learning from scratch. In Figure~\ref{fig:walker_qualitative} we show the behaviors deployed by \textsc{cascade} and \textsc{pp2e} in a single deployment. The  \textsc{cascade} explorers all display diverse, useful behaviors, while \textsc{pp2e} agents are homogenous.

\begin{figure}[h!]
    \centering
    \includegraphics[width=0.95\textwidth]{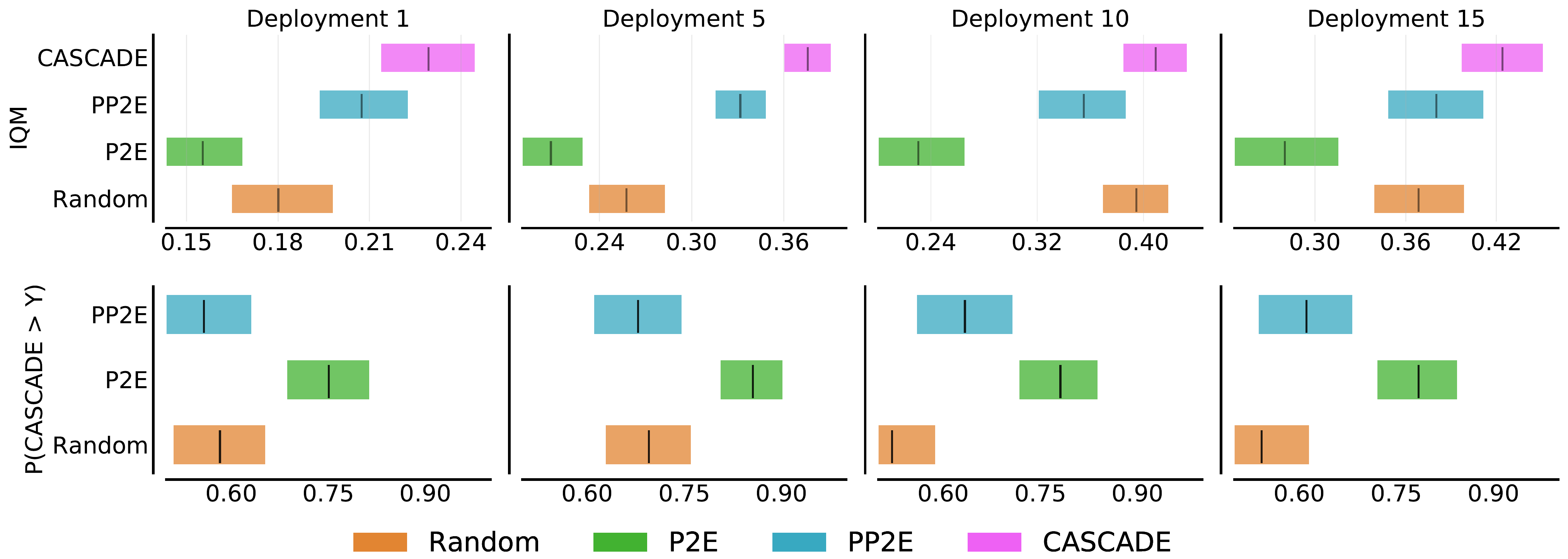}
    \vspace{-1mm}
    \caption{\small{\textbf{DMC Zero-Shot Aggregate Results}: Plots show RLiable metric performance aggregated across all three initial datasets (\texttt{random}, \texttt{medium}, \texttt{expert}) for a total of 30 seeds. Note that the `Probability of Improvement' lower CIs all exceed 0.5, indicating \textsc{cascade} provides a statistically significant improvement (under the Mann-Whitney U test) over all baseline methods \citep{agarwal2021deep}, even after a single deployment.}}
    \vspace{-3mm}
    \label{fig:dmc_rliable}
\end{figure}

We then test the generality of the world model after each deployment by training a separate agent from scratch for each of the four individual tasks: \texttt{stand}, \texttt{walk}, \texttt{run}, \texttt{flip} proposed in URLB \citep{laskin2021urlb}. This demonstrates whether the world model is capable of producing an imaginary MDP that facilitates learning specific behaviors. To compare the performance of \textsc{cascade} and the baselines, in Figure~\ref{fig:dmc_rliable} we show the aggregated statistics from the RLiable library \citep{agarwal2021deep} for a given number of deployments. To do this, we combine the results across all datasets (\texttt{random}, \texttt{medium}, \texttt{expert}) and all downstream tasks (\texttt{stand}, \texttt{walk}, \texttt{run}, \texttt{flip}) and compute the following: 1) the Inter-Quartile Mean (IQM) of the normalized scores (the robust statistic recommended in \citep{agarwal2021deep});  2) the Probability of Improvement, the likelihood of one method outperforming another on a new, unseen task. As we see, \textsc{cascade} shows clear and consistent gains over the next best baseline (\textsc{pp2e}) throughout 15 deployments, showing statistical significance under the Mann-Whitney U test \citep{mannwhitneyu}. 
Note that here the single agent \textsc{P2E} performs poorly, likely due to only collecting data with a single mode of exploratory behavior. See Figure~\ref{fig:dmc_all} in the Appendix for a full breakdown of the performance.

\subsection{Discussion and Limitations}
\label{sec:limitations}

We have shown that \textsc{cascade} is the most effective method in training generally capable agents across a wide range of environments. Indeed, \textsc{cascade} provides the best of both worlds: deep exploration by seeking the frontier of the current knowledge (via InfoGain), while covering the space of possible behaviors (via PopDiv). Each of our baselines lacked in one of these two key properties. Random exploration fared poorly when exploring worlds, which requires deep exploration. However, it can be a highly effective method for exploring behaviors. This is likely why boosting entropy in the action space is so effective for robotics tasks \citep{sac-v2}, but less so in deep exploration tasks \cite{ODonoghue2020Making}. Furthermore, the information theoretic baselines were effective in Exploring Worlds, but the disappointing performance for \textsc{p2e} in DMC is likely caused by a lack of data diversity, while the \textsc{pp2e} populations were not sufficiently heterogeneous to match the coverage of \textsc{cascade}. 

In terms of limitations, the main weakness in current methods is the absolute level of performance. Despite outperforming baselines in DMC, the policies learned by \textsc{cascade} are far from optimal. This is likely due to the challenge of the setting (reward-free deployment efficiency) and we believe that future work in the community will be crucial to further improve these methods. Finally, \textsc{cascade} does require training additional agents that may provide additional computational expense. In this setting we assume the time between deployments is arbitrary, but this may not be the case in all settings and future work could consider ways to make this process more efficient and scalable.

\section{Related Work}

At the core of our paper is the notion of training generalist agents, which have recently been of increased interest in RL \citep{gato, lexa2021, xland, Touati2021}. In particular, we focus on the \emph{reward-free} paradigm, since in many cases it may not be possible to know a priori all possible tasks we may wish our agent to solve \citep{jin2020reward}. With this in mind, there has been a surge of interest in methods that can learn in a self-supervised fashion, to build representations that lead to fast (or even zero-shot) adaptation to future tasks. These methods range from competence-based (i.e. learning ``skills'') \citep{valor2018, gregor17, eysenbach2018diversity, smm2019, aps2021, strouse2022learning} to data-based that aim to maximize data diversity \citep{apt21,protorl}.

In this work we take inspiration from \emph{knowledge-based} approaches, which typically seek to learn a world model \citep{rp1, icm, pathak19a, filos2021model, maximumentropymbrl, burda2018exploration, Sharma2020DynamicsAware}. In particular we build on Plan2Explore \citep{sekar2020planning} which trains an exploration agent by maximizing an objective resembling information gain inside the world model. Plan2Explore was the first work that showed it is possible to transfer a model learned without rewards in a zero-shot fashion. However, it differed from our work in that Plan2Explore solely considers the online RL setting, deploying a new exploration policy every timestep. Other recent work examined the effectiveness of reward-free RL for collecting datasets for subsequent offline RL \citep{offlinedatayarats, offlinedatalambert}. Once again however, this work trains the exploration agent online. By contrast, we only deploy a handful of exploration agents and collect a larger quantity of data with each one.

Our setting resembles a reward-free version of the \emph{deployment efficiency} paradigm, first introduced in \citep{matsushima2021deploymentefficient}. By contrast, their method focuses on policy optimization in the \emph{supervised} RL paradigm, whereas we focus on exploration and zero-shot generalization in the unsupervised RL setting. Follow up work has remained focused on supervised policy optimization \citep{su2022muro}, whereas we are focused on training generalist agents in the reward-free paradigm. Recent work has also shown the importance of frequent policy updates in ensuring diverse exploration \citep{policychurn}, further supporting the need for a diverse population of explorers when operating in the deployment-efficient setting.

To address the challenges of this novel problem setting, we draw inspiration from a variety of fields. In particular, several works have considered exploration with a population of agents \citep{horde, coordinated_exp, coordinated_exp2}, but these have only considered task-specific online RL. Our choice of a diverse population of agents takes inspiration from the field of Quality Diversity (QD, \citep{qdnature, Cully2015RobotsTC}), where it has been shown that diversity can boost exploration in RL \citep{novelty}. Recently, \citep{daqd} showed it is possible to train diverse populations inside a world model to acquire new skills. As far as we are aware, we are the first to consider a cascading objective for training diverse populations, which is typically done either synchronously \citep{dvd} or sequentially \citep{zhou2022continuously, zahavy2021diversity}.

Also related are works that specifically train policies that are able to maximally cover the state space of a given MDP \cite{flambe, modi2021modelfreerep}. However, these works focus on the more theoretically amenable (and less general) low-rank MDP setting, and hence do not demonstrate any experiments on larger scale problems. Furthermore, we aim to achieve state-space coverage by encouraging \emph{policy diversity}, which is distinct from the approaches taken in these works. We make use of an information theoretic objective, which has been shown to be highly effective for exploration in RL \citep{shakir2015, houthooft2016vime, maxenthazan}. Most similar to our work \citep{renyi2021} consider the reward free RL setting, and show that maximizing entropy can lead to improved exploration that can facilitate transfer to arbitrary reward functions. By contrast we consider the limited deployment setting and use model-based RL. Finally, our work relates to collective intelligence \cite{hatang_collective, collectivegps2017}, since each member of the population contributes a small piece to the puzzle that produces a general world model.

\section{Conclusion}

We introduced the \emph{reward-free deployment efficiency} paradigm---an important problem setting for learning generally capable agents in a scalable fashion. To address the challenges of this setting, we proposed \textsc{cascade}, a theoretically motivated approach that leverages a diverse population of self-supervised exploration agents. We showed in a variety of experiments that \textsc{cascade} produces \emph{general exploration strategies}, adept at both deep exploration \emph{and} gathering sufficiently hetereogeneous datasets that facilitate the learning of diverse downstream behaviors. We believe this work provides a new approach for collecting large, diverse datasets in a self-supervised fashion, inducing a potentially open-ended process for training general world models at scale. 

For future work, we could consider alternative means for computing diversity, for example using optimization landscapes \citep{ridge_rider} or explicit behavioral diversity \citep{eysenbach2018diversity,dvd}. These could also be combined with representations that directly incorporate the behaviors of the policy \citep{bharadhwaj2022information} to further improve the design of the ``summary" embedding space. Furthermore, we would be interested in considering transformer models \cite{transformer} for trajectory modelling \cite{chen2021decision, janner2021sequence} that ought to scale more gracefully with larger batches of data \citep{radford2019language}. Recent work has shown that this class of models can compose unseen complex multi-task behaviors at test time from fixed offline datasets \cite{furuta2022generalized}, as well as leverage significantly large and diverse offline datasets to improve behavioral modeling \cite{reid2022wiki}, highly relevant to both aforementioned desiderata for the reward-free deployment efficiency paradigm. 

\section*{Acknowledgments}
The authors would like to thank Cong Lu for help with Offline DreamerV2 code, and Minqi Jiang and Robert Kirk for useful discussions. This work was funded by Meta AI. 


\newpage

\bibliographystyle{abbrv}
\bibliography{refs}

\begin{thebibliography}{100}

\bibitem{valor2018}
J.~Achiam, H.~Edwards, D.~Amodei, and P.~Abbeel.
\newblock Variational option discovery algorithms.
\newblock {\em CoRR}, abs/1807.10299, 2018.

\bibitem{flambe}
A.~Agarwal, S.~Kakade, A.~Krishnamurthy, and W.~Sun.
\newblock Flambe: Structural complexity and representation learning of low rank
  mdps.
\newblock In H.~Larochelle, M.~Ranzato, R.~Hadsell, M.~Balcan, and H.~Lin,
  editors, {\em Advances in Neural Information Processing Systems}, volume~33,
  pages 20095--20107. Curran Associates, Inc., 2020.

\bibitem{agarwal2021deep}
R.~Agarwal, M.~Schwarzer, P.~S. Castro, A.~C. Courville, and M.~Bellemare.
\newblock Deep reinforcement learning at the edge of the statistical precipice.
\newblock {\em Advances in Neural Information Processing Systems}, 34, 2021.

\bibitem{agrawal2012analysis}
S.~Agrawal and N.~Goyal.
\newblock Analysis of thompson sampling for the multi-armed bandit problem.
\newblock In {\em Conference on learning theory}, pages 39--1. JMLR Workshop
  and Conference Proceedings, 2012.

\bibitem{saycan2022arxiv}
M.~Ahn, A.~Brohan, N.~Brown, Y.~Chebotar, O.~Cortes, B.~David, C.~Finn,
  K.~Gopalakrishnan, K.~Hausman, A.~Herzog, D.~Ho, J.~Hsu, J.~Ibarz, B.~Ichter,
  A.~Irpan, E.~Jang, R.~J. Ruano, K.~Jeffrey, S.~Jesmonth, N.~Joshi, R.~Julian,
  D.~Kalashnikov, Y.~Kuang, K.-H. Lee, S.~Levine, Y.~Lu, L.~Luu, C.~Parada,
  P.~Pastor, J.~Quiambao, K.~Rao, J.~Rettinghouse, D.~Reyes, P.~Sermanet,
  N.~Sievers, C.~Tan, A.~Toshev, V.~Vanhoucke, F.~Xia, T.~Xiao, P.~Xu, S.~Xu,
  and M.~Yan.
\newblock Do as i can and not as i say: Grounding language in robotic
  affordances.
\newblock In {\em arXiv preprint arXiv:2204.01691}, 2022.

\bibitem{amodei2016faulty}
D.~Amodei and J.~Clark.
\newblock {Faulty Reward Functions in the Wild}.
\newblock \url{https://blog.openai.com/faulty-reward-functions/}, 2016.

\bibitem{anand2022procedural}
A.~Anand, J.~C. Walker, Y.~Li, E.~V{\'e}rtes, J.~Schrittwieser, S.~Ozair,
  T.~Weber, and J.~B. Hamrick.
\newblock Procedural generalization by planning with self-supervised world
  models.
\newblock In {\em International Conference on Learning Representations}, 2022.

\bibitem{argenson2021modelbased}
A.~Argenson and G.~Dulac-Arnold.
\newblock Model-based offline planning.
\newblock In {\em International Conference on Learning Representations}, 2021.

\bibitem{rp1}
P.~Ball, J.~Parker-Holder, A.~Pacchiano, K.~Choromanski, and S.~Roberts.
\newblock Ready policy one: World building through active learning.
\newblock In {\em Proceedings of the 37th International Conference on Machine
  Learning, {ICML}}, 2020.

\bibitem{ball2021augwm}
P.~J. Ball, C.~Lu, J.~Parker-Holder, and S.~Roberts.
\newblock Augmented world models facilitate zero-shot dynamics generalization
  from a single offline environment.
\newblock In {\em Proceedings of the 38th International Conference on Machine
  Learning}, pages 619--629, 2021.

\bibitem{loon}
M.~Bellemare, S.~Candido, P.~Castro, J.~Gong, M.~Machado, S.~Moitra, S.~Ponda,
  and Z.~Wang.
\newblock Autonomous navigation of stratospheric balloons using reinforcement
  learning.
\newblock {\em Nature}, 588:77--82, December 2020.

\bibitem{ale}
M.~G. Bellemare, Y.~Naddaf, J.~Veness, and M.~Bowling.
\newblock The {A}rcade {L}earning {E}nvironment: {A}n {E}valuation {P}latform
  for {G}eneral {A}gents.
\newblock {\em CoRR}, abs/1207.4708, 2012.

\bibitem{dota}
C.~Berner, G.~Brockman, B.~Chan, V.~Cheung, P.~Debiak, C.~Dennison, D.~Farhi,
  Q.~Fischer, S.~Hashme, C.~Hesse, R.~J{\'{o}}zefowicz, S.~Gray, C.~Olsson,
  J.~Pachocki, M.~Petrov, H.~P. de~Oliveira~Pinto, J.~Raiman, T.~Salimans,
  J.~Schlatter, J.~Schneider, S.~Sidor, I.~Sutskever, J.~Tang, F.~Wolski, and
  S.~Zhang.
\newblock Dota 2 with large scale deep reinforcement learning.
\newblock {\em CoRR}, abs/1912.06680, 2019.

\bibitem{bharadhwaj2022information}
H.~Bharadhwaj, M.~Babaeizadeh, D.~Erhan, and S.~Levine.
\newblock Information prioritization through empowerment in visual model-based
  {RL}.
\newblock In {\em International Conference on Learning Representations}, 2022.

\bibitem{burda2018exploration}
Y.~Burda, H.~Edwards, A.~Storkey, and O.~Klimov.
\newblock Exploration by random network distillation.
\newblock In {\em International Conference on Learning Representations}, 2019.

\bibitem{Byravan2019ImaginedVG}
A.~Byravan, J.~T. Springenberg, A.~Abdolmaleki, R.~Hafner, M.~Neunert,
  T.~Lampe, N.~Siegel, N.~M.~O. Heess, and M.~A. Riedmiller.
\newblock Imagined value gradients: Model-based policy optimization with
  transferable latent dynamics models.
\newblock In {\em CoRL}, 2019.

\bibitem{amigo}
A.~Campero, R.~Raileanu, H.~Kuttler, J.~B. Tenenbaum, T.~Rockt{\"a}schel, and
  E.~Grefenstette.
\newblock Learning with {AMIG}o: Adversarially motivated intrinsic goals.
\newblock In {\em International Conference on Learning Representations}, 2021.

\bibitem{chen2021decision}
L.~Chen, K.~Lu, A.~Rajeswaran, K.~Lee, A.~Grover, M.~Laskin, P.~Abbeel,
  A.~Srinivas, and I.~Mordatch.
\newblock Decision transformer: Reinforcement learning via sequence modeling.
\newblock In A.~Beygelzimer, Y.~Dauphin, P.~Liang, and J.~W. Vaughan, editors,
  {\em Advances in Neural Information Processing Systems}, 2021.

\bibitem{gym_minigrid}
M.~Chevalier-Boisvert, L.~Willems, and S.~Pal.
\newblock Minimalistic gridworld environment for {O}pen{AI} {G}ym.
\newblock \url{https://github.com/maximecb/gym-minigrid}, 2018.

\bibitem{pets}
K.~Chua, R.~Calandra, R.~McAllister, and S.~Levine.
\newblock Deep reinforcement learning in a handful of trials using
  probabilistic dynamics models.
\newblock In {\em Advances in Neural Information Processing Systems 31}, pages
  4754--4765, 2018.

\bibitem{clavera2018learning}
I.~Clavera, A.~Nagabandi, S.~Liu, R.~S. Fearing, P.~Abbeel, S.~Levine, and
  C.~Finn.
\newblock Learning to adapt in dynamic, real-world environments through
  meta-reinforcement learning.
\newblock In {\em International Conference on Learning Representations}, 2019.

\bibitem{novelty}
E.~Conti, V.~Madhavan, F.~P. Such, J.~Lehman, K.~O. Stanley, and J.~Clune.
\newblock Improving exploration in {E}volution {S}trategies for deep
  reinforcement learning via a population of novelty-seeking agents.
\newblock In {\em Proceedings of the 32Nd International Conference on Neural
  Information Processing Systems}, pages 5032--5043, USA, 2018. Curran
  Associates Inc.

\bibitem{cui2021klevel}
B.~Cui, H.~Hu, L.~Pineda, and J.~N. Foerster.
\newblock K-level reasoning for zero-shot coordination in hanabi.
\newblock In A.~Beygelzimer, Y.~Dauphin, P.~Liang, and J.~W. Vaughan, editors,
  {\em Advances in Neural Information Processing Systems}, 2021.

\bibitem{Cully2015RobotsTC}
A.~Cully, J.~Clune, D.~Tarapore, and J.-B. Mouret.
\newblock Robots that can adapt like animals.
\newblock {\em Nature}, 521:503--507, 2015.

\bibitem{rl_plasma}
J.~Degrave, F.~Felici, J.~Buchli, M.~Neunert, B.~Tracey, F.~Carpanese,
  T.~Ewalds, R.~Hafner, A.~Abdolmaleki, D.~de~las Casas, C.~Donner, L.~Fritz,
  C.~Galperti, A.~Huber, J.~Keeling, M.~Tsimpoukelli, J.~Kay, A.~Merle, J.-M.
  Moret, S.~Noury, F.~Pesamosca, D.~Pfau, O.~Sauter, C.~Sommariva, S.~Coda,
  B.~Duval, A.~Fasoli, P.~Kohli, K.~Kavukcuoglu, D.~Hassabis, and
  M.~Riedmiller.
\newblock Magnetic control of tokamak plasmas through deep reinforcement
  learning.
\newblock {\em Nature}, 602:414--419, February 2022.

\bibitem{coordinated_exp}
M.~Dimakopoulou, I.~Osband, and B.~Van~Roy.
\newblock Scalable coordinated exploration in concurrent reinforcement
  learning.
\newblock In S.~Bengio, H.~Wallach, H.~Larochelle, K.~Grauman, N.~Cesa-Bianchi,
  and R.~Garnett, editors, {\em Advances in Neural Information Processing
  Systems}, volume~31. Curran Associates, Inc., 2018.

\bibitem{Ecoffet2021FirstRT}
A.~Ecoffet, J.~Huizinga, J.~Lehman, K.~O. Stanley, and J.~Clune.
\newblock First return then explore.
\newblock {\em Nature}, 590 7847:580--586, 2021.

\bibitem{eysenbach2018diversity}
B.~Eysenbach, A.~Gupta, J.~Ibarz, and S.~Levine.
\newblock Diversity is all you need: Learning skills without a reward function.
\newblock In {\em International Conference on Learning Representations}, 2019.

\bibitem{NEURIPS2021_26405399}
M.~Fatemi, T.~W. Killian, J.~Subramanian, and M.~Ghassemi.
\newblock Medical dead-ends and learning to identify high-risk states and
  treatments.
\newblock In M.~Ranzato, A.~Beygelzimer, Y.~Dauphin, P.~Liang, and J.~W.
  Vaughan, editors, {\em Advances in Neural Information Processing Systems},
  volume~34, pages 4856--4870. Curran Associates, Inc., 2021.

\bibitem{filos2021model}
A.~Filos, E.~V{\'e}rtes, Z.~Marinho, G.~Farquhar, D.~Borsa, A.~Friesen,
  F.~Behbahani, T.~Schaul, A.~Barreto, and S.~Osindero.
\newblock Model-value inconsistency as a signal for epistemic uncertainty.
\newblock {\em arXiv preprint arXiv:2112.04153}, 2021.

\bibitem{agac}
Y.~Flet-Berliac, J.~Ferret, O.~Pietquin, P.~Preux, and M.~Geist.
\newblock Adversarially guided actor-critic.
\newblock In {\em International Conference on Learning Representations}, 2021.

\bibitem{furuta2022generalized}
H.~Furuta, Y.~Matsuo, and S.~S. Gu.
\newblock Generalized decision transformer for offline hindsight information
  matching.
\newblock In {\em International Conference on Learning Representations}, 2022.

\bibitem{gregor17}
K.~Gregor, D.~J. Rezende, and D.~Wierstra.
\newblock Variational intrinsic control.
\newblock In {\em 5th International Conference on Learning Representations,
  {ICLR} 2017, Toulon, France, April 24-26, 2017, Workshop Track Proceedings},
  2017.

\bibitem{worldmodels}
D.~Ha and J.~Schmidhuber.
\newblock Recurrent world models facilitate policy evolution.
\newblock In {\em Proceedings of the 32Nd International Conference on Neural
  Information Processing Systems}, NeurIPS'18, pages 2455--2467, 2018.

\bibitem{hatang_collective}
D.~Ha and Y.~Tang.
\newblock Collective intelligence for deep learning: {A} survey of recent
  developments.
\newblock {\em CoRR}, abs/2111.14377, 2021.

\bibitem{sac-v2}
T.~Haarnoja, A.~Zhou, K.~Hartikainen, G.~Tucker, S.~Ha, J.~Tan, V.~Kumar,
  H.~Zhu, A.~Gupta, P.~Abbeel, and S.~Levine.
\newblock Soft actor-critic algorithms and applications.
\newblock {\em CoRR}, abs/1812.05905, 2018.

\bibitem{hafner2022benchmarking}
D.~Hafner.
\newblock Benchmarking the spectrum of agent capabilities.
\newblock In {\em International Conference on Learning Representations}, 2022.

\bibitem{dreamer}
D.~Hafner, T.~Lillicrap, J.~Ba, and M.~Norouzi.
\newblock Dream to control: Learning behaviors by latent imagination.
\newblock In {\em International Conference on Learning Representations}, 2020.

\bibitem{planet}
D.~Hafner, T.~Lillicrap, I.~Fischer, R.~Villegas, D.~Ha, H.~Lee, and
  J.~Davidson.
\newblock Learning latent dynamics for planning from pixels.
\newblock In K.~Chaudhuri and R.~Salakhutdinov, editors, {\em Proceedings of
  the 36th International Conference on Machine Learning}, pages 2555--2565,
  2019.

\bibitem{hafner2021mastering}
D.~Hafner, T.~P. Lillicrap, M.~Norouzi, and J.~Ba.
\newblock Mastering atari with discrete world models.
\newblock In {\em International Conference on Learning Representations}, 2021.

\bibitem{maxenthazan}
E.~Hazan, S.~Kakade, K.~Singh, and A.~Van~Soest.
\newblock Provably efficient maximum entropy exploration.
\newblock In K.~Chaudhuri and R.~Salakhutdinov, editors, {\em Proceedings of
  the 36th International Conference on Machine Learning}, volume~97 of {\em
  Proceedings of Machine Learning Research}, pages 2681--2691. PMLR, 09--15 Jun
  2019.

\bibitem{bald}
N.~Houlsby, F.~Huszar, Z.~Ghahramani, and M.~Lengyel.
\newblock Bayesian active learning for classification and preference learning.
\newblock {\em CoRR}, abs/1112.5745, 2011.

\bibitem{houthooft2016vime}
R.~Houthooft, X.~Chen, Y.~Duan, J.~Schulman, F.~De~Turck, and P.~Abbeel.
\newblock Vime: Variational information maximizing exploration.
\newblock {\em Advances in neural information processing systems}, 29, 2016.

\bibitem{janner2021sequence}
M.~Janner, Q.~Li, and S.~Levine.
\newblock Offline reinforcement learning as one big sequence modeling problem.
\newblock In {\em Advances in Neural Information Processing Systems}, 2021.

\bibitem{jin2020reward}
C.~Jin, A.~Krishnamurthy, M.~Simchowitz, and T.~Yu.
\newblock Reward-free exploration for reinforcement learning.
\newblock In {\em International Conference on Machine Learning}, pages
  4870--4879. PMLR, 2020.

\bibitem{pcg_illuminating}
N.~Justesen, R.~R. Torrado, P.~Bontrager, A.~Khalifa, J.~Togelius, and S.~Risi.
\newblock Procedural level generation improves generality of deep reinforcement
  learning.
\newblock {\em CoRR}, abs/1806.10729, 2018.

\bibitem{simple}
L.~Kaiser, M.~Babaeizadeh, P.~Milos, B.~Osiński, R.~H. Campbell,
  K.~Czechowski, D.~Erhan, C.~Finn, P.~Kozakowski, S.~Levine, A.~Mohiuddin,
  R.~Sepassi, G.~Tucker, and H.~Michalewski.
\newblock Model based reinforcement learning for {A}tari.
\newblock In {\em International Conference on Learning Representations}, 2020.

\bibitem{kalashnikov2018scalable}
D.~Kalashnikov, A.~Irpan, P.~Pastor, J.~Ibarz, A.~Herzog, E.~Jang, D.~Quillen,
  E.~Holly, M.~Kalakrishnan, V.~Vanhoucke, et~al.
\newblock Scalable deep reinforcement learning for vision-based robotic
  manipulation.
\newblock In {\em Conference on Robot Learning}, pages 651--673. PMLR, 2018.

\bibitem{morel}
R.~Kidambi, A.~Rajeswaran, P.~Netrapalli, and T.~Joachims.
\newblock Morel : Model-based offline reinforcement learning.
\newblock In {\em Advances in Neural Information Processing Systems}, 2020.

\bibitem{kingma2013auto}
D.~P. Kingma and M.~Welling.
\newblock Auto-encoding variational bayes.
\newblock In {\em ICLR}, 2014.

\bibitem{kirk2021survey}
R.~Kirk, A.~Zhang, E.~Grefenstette, and T.~Rockt{\"a}schel.
\newblock A survey of generalisation in deep reinforcement learning.
\newblock {\em arXiv preprint arXiv:2111.09794}, 2021.

\bibitem{kirsch2019batchbald}
A.~Kirsch, J.~Van~Amersfoort, and Y.~Gal.
\newblock Batchbald: Efficient and diverse batch acquisition for deep bayesian
  active learning.
\newblock {\em Advances in neural information processing systems}, 32, 2019.

\bibitem{KoberRLRoboticsSurvey2013}
J.~Kober, J.~A. Bagnell, and J.~Peters.
\newblock Reinforcement learning in robotics: A survey.
\newblock {\em The International Journal of Robotics Research},
  32(11):1238--1274, 2013.

\bibitem{METRPO}
T.~Kurutach, I.~Clavera, Y.~Duan, A.~Tamar, and P.~Abbeel.
\newblock Model-ensemble trust-region policy optimization.
\newblock In {\em International Conference on Learning Representations}, 2018.

\bibitem{offlinedatalambert}
N.~Lambert, M.~Wulfmeier, W.~F. Whitney, A.~Byravan, M.~Bloesch, V.~Dasagi,
  T.~Hertweck, and M.~A. Riedmiller.
\newblock The challenges of exploration for offline reinforcement learning.
\newblock {\em CoRR}, 2022.

\bibitem{laskin2021urlb}
M.~Laskin, D.~Yarats, H.~Liu, K.~Lee, A.~Zhan, K.~Lu, C.~Cang, L.~Pinto, and
  P.~Abbeel.
\newblock {URLB}: Unsupervised reinforcement learning benchmark.
\newblock In {\em Thirty-fifth Conference on Neural Information Processing
  Systems Datasets and Benchmarks Track (Round 2)}, 2021.

\bibitem{cnn}
Y.~LeCun, B.~Boser, J.~S. Denker, D.~Henderson, R.~E. Howard, W.~Hubbard, and
  L.~D. Jackel.
\newblock Backpropagation applied to handwritten zip code recognition.
\newblock In {\em Neural Computation}, volume~1, pages 541--551, 1989.

\bibitem{lee2020context}
K.~Lee, Y.~Seo, S.~Lee, H.~Lee, and J.~Shin.
\newblock Context-aware dynamics model for generalization in model-based
  reinforcement learning.
\newblock In {\em International Conference on Machine Learning}, 2020.

\bibitem{smm2019}
L.~Lee, B.~Eysenbach, E.~Parisotto, E.~Xing, S.~Levine, and R.~Salakhutdinov.
\newblock Efficient exploration via state marginal matching, 2019.

\bibitem{armfarm}
S.~Levine, P.~Pastor, A.~Krizhevsky, J.~Ibarz, and D.~Quillen.
\newblock Learning hand-eye coordination for robotic grasping with deep
  learning and large-scale data collection.
\newblock {\em Int. J. Robotics Res.}, 37(4-5):421--436, 2018.

\bibitem{daqd}
B.~Lim, L.~Grillotti, L.~Bernasconi, and A.~Cully.
\newblock Dynamics-aware quality-diversity for efficient learning of skill
  repertoires.
\newblock In {\em ICRA}, 2022.

\bibitem{aps2021}
H.~Liu and P.~Abbeel.
\newblock Aps: Active pretraining with successor features.
\newblock In {\em Proceedings of the 38th International Conference on Machine
  Learning}, pages 6736--6747, 2021.

\bibitem{apt21}
H.~Liu and P.~Abbeel.
\newblock Behavior from the void: Unsupervised active pre-training, 2021.

\bibitem{coordinated_exp2}
I.-J. Liu, U.~Jain, R.~A. Yeh, and A.~Schwing.
\newblock Cooperative exploration for multi-agent deep reinforcement learning.
\newblock In M.~Meila and T.~Zhang, editors, {\em Proceedings of the 38th
  International Conference on Machine Learning}, volume 139 of {\em Proceedings
  of Machine Learning Research}, pages 6826--6836. PMLR, 18--24 Jul 2021.

\bibitem{education}
T.~Mandel, Y.-E. Liu, S.~Levine, E.~Brunskill, and Z.~Popovic.
\newblock Offline policy evaluation across representations with applications to
  educational games.
\newblock In {\em Proceedings of the 2014 International Conference on
  Autonomous Agents and Multi-Agent Systems}, AAMAS '14, page 1077–1084,
  Richland, SC, 2014. International Foundation for Autonomous Agents and
  Multiagent Systems.

\bibitem{mannwhitneyu}
H.~B. Mann and D.~R. Whitney.
\newblock {On a Test of Whether one of Two Random Variables is Stochastically
  Larger than the Other}.
\newblock {\em The Annals of Mathematical Statistics}, 18(1):50 -- 60, 1947.

\bibitem{matsushima2021deploymentefficient}
T.~Matsushima, H.~Furuta, Y.~Matsuo, O.~Nachum, and S.~Gu.
\newblock Deployment-efficient reinforcement learning via model-based offline
  optimization.
\newblock In {\em International Conference on Learning Representations}, 2021.

\bibitem{lexa2021}
R.~Mendonca, O.~Rybkin, K.~Daniilidis, D.~Hafner, and D.~Pathak.
\newblock Discovering and achieving goals via world models, 2021.

\bibitem{dqn}
V.~Mnih, K.~Kavukcuoglu, D.~Silver, A.~Rusu, J.~Veness, M.~Bellemare,
  A.~Graves, M.~Riedmiller, A.~Fidjeland, G.~Ostrovski, S.~Petersen,
  C.~Beattie, A.~Sadik, I.~Antonoglou, H.~King, D.~Kumaran, D.~Wierstra,
  S.~Legg, and D.~Hassabis.
\newblock Human-level control through deep reinforcement learning.
\newblock {\em Nature}, 518:529--33, 02 2015.

\bibitem{modi2021modelfreerep}
A.~Modi, J.~Chen, A.~Krishnamurthy, N.~Jiang, and A.~Agarwal.
\newblock Model-free representation learning and exploration in low-rank mdps,
  2021.

\bibitem{shakir2015}
S.~Mohamed and D.~J. Rezende.
\newblock Variational information maximisation for intrinsically motivated
  reinforcement learning.
\newblock In {\em Proceedings of the 28th International Conference on Neural
  Information Processing Systems - Volume 2}, NIPS'15, page 2125–2133,
  Cambridge, MA, USA, 2015. MIT Press.

\bibitem{moskovitz2021efficient}
T.~Moskovitz, M.~Arbel, F.~Huszar, and A.~Gretton.
\newblock Efficient wasserstein natural gradients for reinforcement learning.
\newblock In {\em International Conference on Learning Representations}, 2021.

\bibitem{nemhauser1978analysis}
G.~L. Nemhauser, L.~A. Wolsey, and M.~L. Fisher.
\newblock An analysis of approximations for maximizing submodular set
  functions—i.
\newblock {\em Mathematical programming}, 14(1):265--294, 1978.

\bibitem{Ng99policyinvariance}
A.~Y. Ng, D.~Harada, and S.~Russell.
\newblock Policy invariance under reward transformations: Theory and
  application to reward shaping.
\newblock In {\em In Proceedings of the Sixteenth International Conference on
  Machine Learning}, pages 278--287. Morgan Kaufmann, 1999.

\bibitem{ODonoghue2020Making}
B.~O'Donoghue, I.~Osband, and C.~Ionescu.
\newblock Making sense of reinforcement learning and probabilistic inference.
\newblock In {\em International Conference on Learning Representations}, 2020.

\bibitem{oh2015}
J.~Oh, X.~Guo, H.~Lee, R.~Lewis, and S.~Singh.
\newblock Action-conditional video prediction using deep networks in atari
  games.
\newblock In {\em Proceedings of the 28th International Conference on Neural
  Information Processing Systems - Volume 2}, NIPS'15, page 2863–2871,
  Cambridge, MA, USA, 2015. MIT Press.

\bibitem{dexterous_openai}
OpenAI, M.~Andrychowicz, B.~Baker, M.~Chociej, R.~J{\'{o}}zefowicz, B.~McGrew,
  J.~W. Pachocki, J.~Pachocki, A.~Petron, M.~Plappert, G.~Powell, A.~Ray,
  J.~Schneider, S.~Sidor, J.~Tobin, P.~Welinder, L.~Weng, and W.~Zaremba.
\newblock Learning dexterous in-hand manipulation.
\newblock {\em CoRR}, abs/1808.00177, 2018.

\bibitem{osband2019deepexplore}
I.~Osband, B.~V. Roy, D.~J. Russo, and Z.~Wen.
\newblock Deep exploration via randomized value functions.
\newblock {\em Journal of Machine Learning Research}, 20(124):1--62, 2019.

\bibitem{wass}
A.~Pacchiano, J.~Parker-Holder, Y.~Tang, A.~Choromanska, K.~Choromanski, and
  M.~I. Jordan.
\newblock Learning to score behaviors for guided policy optimization.
\newblock In {\em Proceedings of the 37th International Conference on Machine
  Learning}, 2020.

\bibitem{packer2018assessing}
C.~Packer, K.~Gao, J.~Kos, P.~Kr{\"a}henb{\"u}hl, V.~Koltun, and D.~Song.
\newblock Assessing generalization in deep reinforcement learning.
\newblock {\em arXiv preprint arXiv:1810.12282}, 2018.

\bibitem{ridge_rider}
J.~Parker-Holder, L.~Metz, C.~Resnick, H.~Hu, A.~Lerer, A.~Letcher,
  A.~Peysakhovich, A.~Pacchiano, and J.~Foerster.
\newblock Ridge rider: Finding diverse solutions by following eigenvectors of
  the hessian.
\newblock In H.~Larochelle, M.~Ranzato, R.~Hadsell, M.~Balcan, and H.~Lin,
  editors, {\em Advances in Neural Information Processing Systems}, volume~33,
  pages 753--765. Curran Associates, Inc., 2020.

\bibitem{dvd}
J.~Parker{-}Holder, A.~Pacchiano, K.~Choromanski, and S.~Roberts.
\newblock Effective diversity in population-based reinforcement learning.
\newblock In {\em Advances in Neural Information Processing Systems 33}, 2020.

\bibitem{icm}
D.~Pathak, P.~Agrawal, A.~A. Efros, and T.~Darrell.
\newblock Curiosity-driven exploration by self-supervised prediction.
\newblock In {\em Proceedings of the 34th International Conference on Machine
  Learning - Volume 70}, ICML'17, page 2778–2787. JMLR.org, 2017.

\bibitem{pathak19a}
D.~Pathak, D.~Gandhi, and A.~Gupta.
\newblock Self-supervised exploration via disagreement.
\newblock In K.~Chaudhuri and R.~Salakhutdinov, editors, {\em Proceedings of
  the 36th International Conference on Machine Learning}, volume~97 of {\em
  Proceedings of Machine Learning Research}, pages 5062--5071. PMLR, 09--15 Jun
  2019.

\bibitem{qdnature}
J.~K. Pugh, L.~B. Soros, and K.~O. Stanley.
\newblock Quality diversity: A new frontier for evolutionary computation.
\newblock {\em Frontiers in Robotics and AI}, 3:40, 2016.

\bibitem{qiu2021reward}
S.~Qiu, J.~Ye, Z.~Wang, and Z.~Yang.
\newblock On reward-free rl with kernel and neural function approximations:
  Single-agent mdp and markov game.
\newblock In {\em International Conference on Machine Learning}, pages
  8737--8747. PMLR, 2021.

\bibitem{radford2019language}
A.~Radford, J.~Wu, R.~Child, D.~Luan, D.~Amodei, and I.~Sutskever.
\newblock Language models are unsupervised multitask learners.
\newblock 2019.

\bibitem{lompo}
R.~Rafailov, T.~Yu, A.~Rajeswaran, and C.~Finn.
\newblock Offline reinforcement learning from images with latent space models.
\newblock In {\em Proceedings of the 3rd Conference on Learning for Dynamics
  and Control}, volume 144 of {\em Proceedings of Machine Learning Research},
  pages 1154--1168. PMLR, 2021.

\bibitem{ride}
R.~Raileanu and T.~Rocktäschel.
\newblock Ride: Rewarding impact-driven exploration for procedurally-generated
  environments.
\newblock In {\em International Conference on Learning Representations}, 2020.

\bibitem{gato}
S.~Reed, K.~Zolna, E.~Parisotto, S.~G. Colmenarejo, A.~Novikov, G.~Barth-Maron,
  M.~Gimenez, Y.~Sulsky, J.~Kay, J.~T. Springenberg, T.~Eccles, J.~Bruce,
  A.~Razavi, A.~Edwards, N.~Heess, Y.~Chen, R.~Hadsell, O.~Vinyals, M.~Bordbar,
  and N.~de~Freitas.
\newblock A generalist agent, 2022.

\bibitem{reid2022wiki}
M.~Reid, Y.~Yamada, and S.~S. Gu.
\newblock Can wikipedia help offline reinforcement learning?, 2022.

\bibitem{rezende2014stochastic}
D.~J. Rezende, S.~Mohamed, and D.~Wierstra.
\newblock Stochastic backpropagation and approximate inference in deep
  generative models.
\newblock In {\em International conference on machine learning}, pages
  1278--1286. PMLR, 2014.

\bibitem{pcg}
S.~Risi and J.~Togelius.
\newblock Increasing generality in machine learning through procedural content
  generation.
\newblock {\em Nature Machine Intelligence}, 2, 08 2020.

\bibitem{rnn}
D.~E. {Rumelhart}, G.~E. {Hinton}, and R.~J. {Williams}.
\newblock {Learning representations by back-propagating errors}.
\newblock {\em Nature}, 323(6088):533--536, Oct. 1986.

\bibitem{policychurn}
T.~Schaul, A.~Barreto, J.~Quan, and G.~Ostrovski.
\newblock The phenomenon of policy churn, 2022.

\bibitem{Schmidhuber90makingthe}
J.~Schmidhuber.
\newblock Making the world differentiable: On using self-supervised fully
  recurrent neural networks for dynamic reinforcement learning and planning in
  non-stationary environments.
\newblock Technical report, 1990.

\bibitem{sekar2020planning}
R.~Sekar, O.~Rybkin, K.~Daniilidis, P.~Abbeel, D.~Hafner, and D.~Pathak.
\newblock Planning to explore via self-supervised world models.
\newblock In {\em International Conference on Machine Learning}, pages
  8583--8592. PMLR, 2020.

\bibitem{Sharma2020DynamicsAware}
A.~Sharma, S.~Gu, S.~Levine, V.~Kumar, and K.~Hausman.
\newblock Dynamics-aware unsupervised discovery of skills.
\newblock In {\em International Conference on Learning Representations}, 2020.

\bibitem{shyam2019model}
P.~Shyam, W.~Ja{\'s}kowski, and F.~Gomez.
\newblock Model-based active exploration.
\newblock In {\em International conference on machine learning}, pages
  5779--5788. PMLR, 2019.

\bibitem{alphago}
D.~Silver, A.~Huang, C.~J. Maddison, A.~Guez, L.~Sifre, G.~van~den Driessche,
  J.~Schrittwieser, I.~Antonoglou, V.~Panneershelvam, M.~Lanctot, S.~Dieleman,
  D.~Grewe, J.~Nham, N.~Kalchbrenner, I.~Sutskever, T.~P. Lillicrap, M.~Leach,
  K.~Kavukcuoglu, T.~Graepel, and D.~Hassabis.
\newblock Mastering the game of {G}o with deep neural networks and tree search.
\newblock {\em Nature}, 529:484--489, 2016.

\bibitem{strouse2022learning}
D.~Strouse, K.~Baumli, D.~Warde-Farley, V.~Mnih, and S.~S. Hansen.
\newblock Learning more skills through optimistic exploration.
\newblock In {\em International Conference on Learning Representations}, 2022.

\bibitem{su2022muro}
D.~Su, J.~D. Lee, J.~Mulvey, and H.~V. Poor.
\newblock {MURO}: Deployment constrained reinforcement learning with
  model-based uncertainty regularized batch optimization, 2022.

\bibitem{seq2seq}
I.~Sutskever, O.~Vinyals, and Q.~V. Le.
\newblock Sequence to sequence learning with neural networks.
\newblock In Z.~Ghahramani, M.~Welling, C.~Cortes, N.~Lawrence, and
  K.~Weinberger, editors, {\em Advances in Neural Information Processing
  Systems}, volume~27. Curran Associates, Inc., 2014.

\bibitem{dyna}
R.~S. Sutton.
\newblock Dyna, an integrated architecture for learning, planning, and
  reacting.
\newblock {\em SIGART Bull.}, 2(4):160–163, July 1991.

\bibitem{Sutton1998}
R.~S. Sutton and A.~G. Barto.
\newblock {\em Introduction to Reinforcement Learning}.
\newblock MIT Press, Cambridge, MA, USA, 1st edition, 1998.

\bibitem{horde}
R.~S. Sutton, J.~Modayil, M.~Delp, T.~Degris, P.~M. Pilarski, A.~White, and
  D.~Precup.
\newblock Horde: A scalable real-time architecture for learning knowledge from
  unsupervised sensorimotor interaction.
\newblock In {\em The 10th International Conference on Autonomous Agents and
  Multiagent Systems - Volume 2}, AAMAS '11, page 761–768, Richland, SC,
  2011. International Foundation for Autonomous Agents and Multiagent Systems.

\bibitem{maximumentropymbrl}
O.~Svidchenko and A.~Shpilman.
\newblock Maximum entropy model-based reinforcement learning.
\newblock {\em CoRR}, abs/2112.01195, 2021.

\bibitem{MingCostSens1993}
M.~Tan.
\newblock Cost-sensitive learning of classification knowledge and its
  applications in robotics.
\newblock {\em Mach. Learn.}, 13(1):7–33, oct 1993.

\bibitem{xland}
O.~E.~L. Team, A.~Stooke, A.~Mahajan, C.~Barros, C.~Deck, J.~Bauer,
  J.~Sygnowski, M.~Trebacz, M.~Jaderberg, M.~Mathieu, N.~McAleese,
  N.~Bradley{-}Schmieg, N.~Wong, N.~Porcel, R.~Raileanu, S.~Hughes{-}Fitt,
  V.~Dalibard, and W.~M. Czarnecki.
\newblock Open-ended learning leads to generally capable agents.
\newblock {\em CoRR}, abs/2107.12808, 2021.

\bibitem{Touati2021}
A.~Touati and Y.~Ollivier.
\newblock Learning one representation to optimize all rewards.
\newblock In {\em Advances in Neural Information Processing Systems},
  volume~34, pages 13--23, 2021.

\bibitem{tunyasuvunakool2020}
S.~Tunyasuvunakool, A.~Muldal, Y.~Doron, S.~Liu, S.~Bohez, J.~Merel, T.~Erez,
  T.~Lillicrap, N.~Heess, and Y.~Tassa.
\newblock dm control: Software and tasks for continuous control.
\newblock {\em Software Impacts}, 6:100022, 2020.

\bibitem{hasselt2019der}
H.~van Hasselt, M.~Hessel, and J.~Aslanides.
\newblock When to use parametric models in reinforcement learning?
\newblock In {\em Advances in Neural Information Processing Systems}, pages
  14322--14333, 2019.

\bibitem{transformer}
A.~Vaswani, N.~Shazeer, N.~Parmar, J.~Uszkoreit, L.~Jones, A.~N. Gomez, L.~u.
  Kaiser, and I.~Polosukhin.
\newblock Attention is all you need.
\newblock In I.~Guyon, U.~V. Luxburg, S.~Bengio, H.~Wallach, R.~Fergus,
  S.~Vishwanathan, and R.~Garnett, editors, {\em Advances in Neural Information
  Processing Systems}, volume~30. Curran Associates, Inc., 2017.

\bibitem{wang2020reward}
R.~Wang, S.~S. Du, L.~Yang, and R.~R. Salakhutdinov.
\newblock On reward-free reinforcement learning with linear function
  approximation.
\newblock {\em Advances in neural information processing systems},
  33:17816--17826, 2020.

\bibitem{collectivegps2017}
A.~Yahya, A.~Li, M.~Kalakrishnan, Y.~Chebotar, and S.~Levine.
\newblock Collective robot reinforcement learning with distributed asynchronous
  guided policy search.
\newblock In {\em 2017 IEEE/RSJ International Conference on Intelligent Robots
  and Systems (IROS)}, pages 79--86, 2017.

\bibitem{offlinedatayarats}
D.~Yarats, D.~Brandfonbrener, H.~Liu, M.~Laskin, P.~Abbeel, A.~Lazaric, and
  L.~Pinto.
\newblock Don't change the algorithm, change the data: Exploratory data for
  offline reinforcement learning.
\newblock {\em CoRR}, abs/2201.13425, 2022.

\bibitem{protorl}
D.~Yarats, R.~Fergus, A.~Lazaric, and L.~Pinto.
\newblock Reinforcement learning with prototypical representations.
\newblock In {\em Proceedings of the 38th International Conference on Machine
  Learning}, pages 11920--11931, 2021.

\bibitem{mopo}
T.~Yu, G.~Thomas, L.~Yu, S.~Ermon, J.~Zou, S.~Levine, C.~Finn, and T.~Ma.
\newblock Mopo: Model-based offline policy optimization.
\newblock In {\em Advances in Neural Information Processing Systems}, 2020.

\bibitem{zahavy2021diversity}
T.~Zahavy, Y.~Schroecker, F.~Behbahani, K.~Baumli, S.~Flennerhag, S.~Hou, and
  S.~Singh.
\newblock Discovering policies with domino: Diversity optimization maintaining
  near optimality.
\newblock {\em arXiv preprint arXiv:2205.13521}, 2022.

\bibitem{renyi2021}
C.~Zhang, Y.~Cai, L.~Huang, and J.~Li.
\newblock Exploration by maximizing renyi entropy for reward-free {RL}
  framework.
\newblock In {\em Thirty-Fifth {AAAI} Conference on Artificial Intelligence,
  {AAAI} 2021, Thirty-Third Conference on Innovative Applications of Artificial
  Intelligence, {IAAI} 2021, The Eleventh Symposium on Educational Advances in
  Artificial Intelligence, {EAAI} 2021, Virtual Event, February 2-9, 2021},
  pages 10859--10867. {AAAI} Press, 2021.

\bibitem{zhang2021noveld}
T.~Zhang, H.~Xu, X.~Wang, Y.~Wu, K.~Keutzer, J.~E. Gonzalez, and Y.~Tian.
\newblock Noveld: A simple yet effective exploration criterion.
\newblock In A.~Beygelzimer, Y.~Dauphin, P.~Liang, and J.~W. Vaughan, editors,
  {\em Advances in Neural Information Processing Systems}, 2021.

\bibitem{zhou2022continuously}
Z.~Zhou, W.~Fu, B.~Zhang, and Y.~Wu.
\newblock Continuously discovering novel strategies via reward-switching policy
  optimization.
\newblock In {\em International Conference on Learning Representations}, 2022.

\end{thebibliography}

\newpage
\section*{Checklist}

\begin{enumerate}

\item For all authors...
\begin{enumerate}
  \item Do the main claims made in the abstract and introduction accurately reflect the paper's contributions and scope?
    \answerYes{}
  \item Did you describe the limitations of your work?
    \answerYes{See Section~\ref{sec:limitations}}
  \item Did you discuss any potential negative societal impacts of your work?
    \answerNA{}
  \item Have you read the ethics review guidelines and ensured that your paper conforms to them?
    \answerYes{}
\end{enumerate}

\item If you are including theoretical results...
\begin{enumerate}
  \item Did you state the full set of assumptions of all theoretical results?
    \answerYes{}
        \item Did you include complete proofs of all theoretical results?
    \answerYes{}
\end{enumerate}

\item If you ran experiments...
\begin{enumerate}
  \item Did you include the code, data, and instructions needed to reproduce the main experimental results (either in the supplemental material or as a URL)?
    \answerYes{}
  \item Did you specify all the training details (e.g., data splits, hyperparameters, how they were chosen)?
    \answerYes{}
        \item Did you report error bars (e.g., with respect to the random seed after running experiments multiple times)?
    \answerYes{}
        \item Did you include the total amount of compute and the type of resources used (e.g., type of GPUs, internal cluster, or cloud provider)?
    \answerYes{}
\end{enumerate}

\item If you are using existing assets (e.g., code, data, models) or curating/releasing new assets...
\begin{enumerate}
  \item If your work uses existing assets, did you cite the creators?
    \answerYes{}
  \item Did you mention the license of the assets?
    \answerNA{}
  \item Did you include any new assets either in the supplemental material or as a URL?
    \answerNo{}
  \item Did you discuss whether and how consent was obtained from people whose data you're using/curating?
    \answerNA{}
  \item Did you discuss whether the data you are using/curating contains personally identifiable information or offensive content?
    \answerNA{}
\end{enumerate}

\item If you used crowdsourcing or conducted research with human subjects...
\begin{enumerate}
  \item Did you include the full text of instructions given to participants and screenshots, if applicable?
    \answerNA{}
  \item Did you describe any potential participant risks, with links to Institutional Review Board (IRB) approvals, if applicable?
    \answerNA{}
  \item Did you include the estimated hourly wage paid to participants and the total amount spent on participant compensation?
    \answerNA{}
\end{enumerate}
\end{enumerate}

\newpage
\appendix

\section*{Appendix}
\section{Additional Experimental Results}

\subsection{DMC}

\begin{figure}[H]
    \centering
    \includegraphics[width=0.9\textwidth]{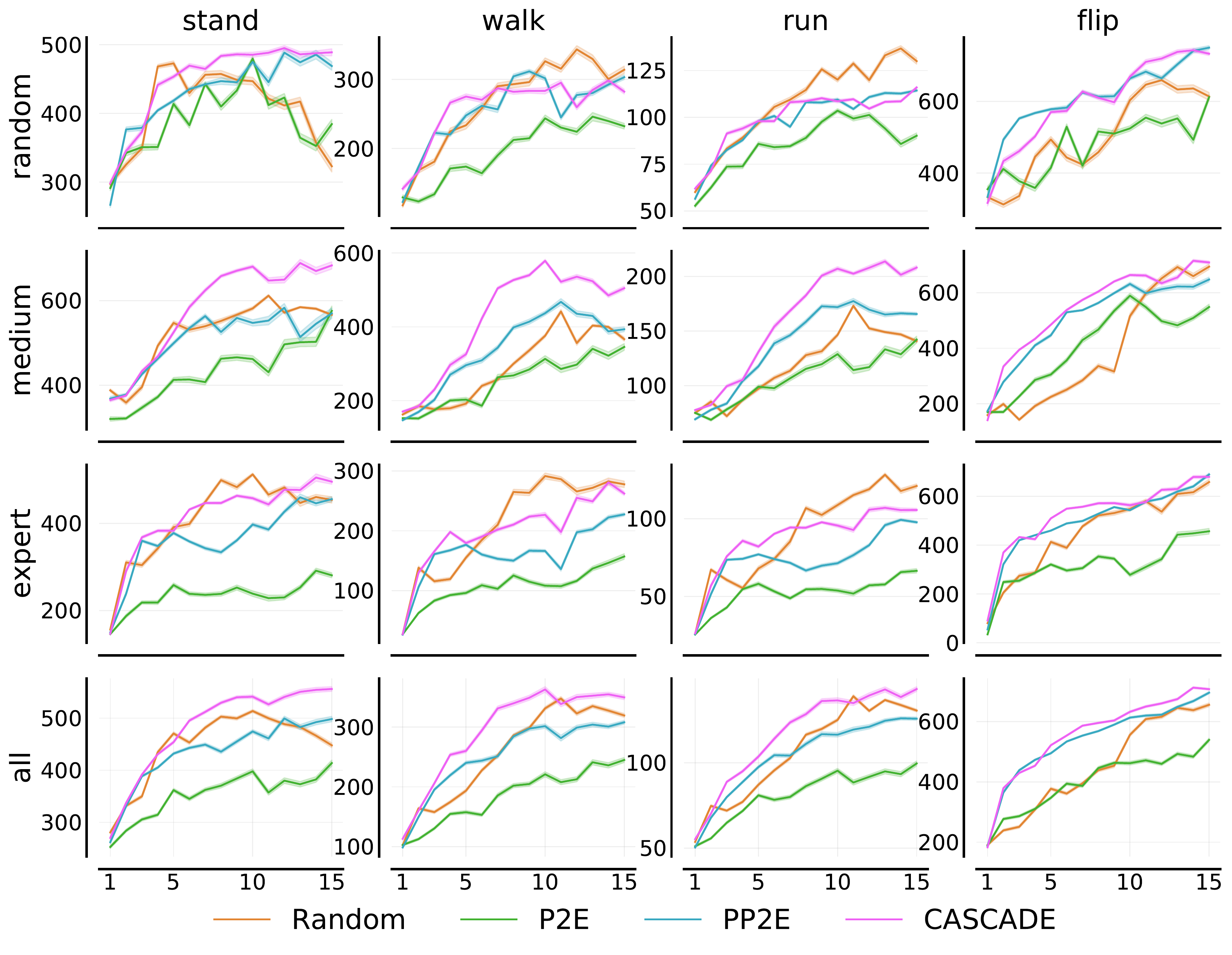}
    \caption{\small{\textbf{DMC Zero-Shot Results}: Plots show the performance by environment and dataset, with the last row being the task performance average over all three initial datasets. Plots show the mean and SEM over 10 seeds.}}
    \label{fig:dmc_all}
\end{figure}

Here we show additional experimental results, beginning with a full breakdown of the DMC results for each dataset. In Figure~\ref{fig:dmc_all} we see the zero-shot transfer performance for models trained with \texttt{random}, \texttt{medium} and \texttt{expert} initial datasets, and 14 subsequent deployments. Interestingly, we see that the \texttt{medium} dataset proves to be the most effective for achieving high performance, this may be due to being more diverse than the \texttt{expert} dataset, which is a narrow distribution of high performing episodes. 

\newpage
\subsection{Atari Tasks}
\label{app:atari_deppdive}

\begin{figure}[H]
    \centering
    \includegraphics[width=0.99\textwidth]{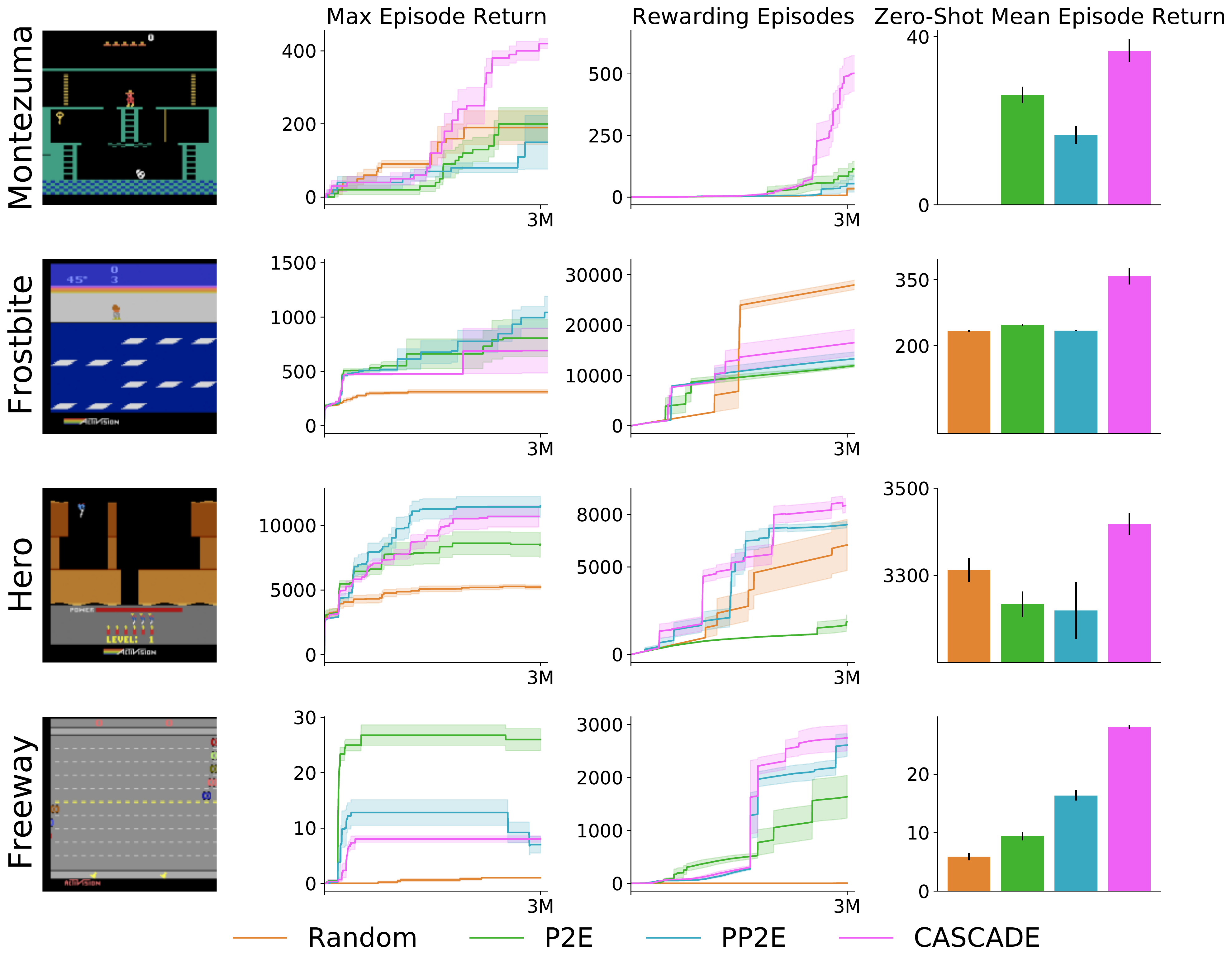}
    \caption{\small{{From left to right on each row we show a frame from the game, a plot of the
cumulative maximum episode reward, the total number of rewarding episodes discovered and zero-shot average
return, from 3M training steps (15 deployments). Plots show mean and SEM over 10 seeds. Note that the max episode returns of all methods stabilize after 15 deployments (3M training steps in total).}}}
    \label{fig:atari_mont_full}
\end{figure}

Next we show detailed results in four Atari games: Montezuma's Revenge, Frostbite, Hero and Freeway in Fig.\ \ref{fig:atari_mont_full}. We see that generally \textsc{cascade} discovers higher rewarding episodes (almost double the next best baseline in Montezuma's Revenge), and also discovers more rewarding episodes on average. This translates to stronger zero-shot performance than all baselines in all four environments. On the other hand, the other baselines achieve worse zero-shot performance because they either (1) find an occasional high reward episode (Max Episode Return), but not in a sufficient quantity (Rewarding Episodes), or (2) collect abundant low rewarding episodes that are less helpful for learning good behavior polices. For instance, the random baseline is able to find Rewarding Episodes more frequently in Frostbite, but these are relatively low quality, hence the lower curve when assessing its Max Episode Return. 

\begin{figure}[H]
    \centering
    \includegraphics[width=0.6\textwidth]{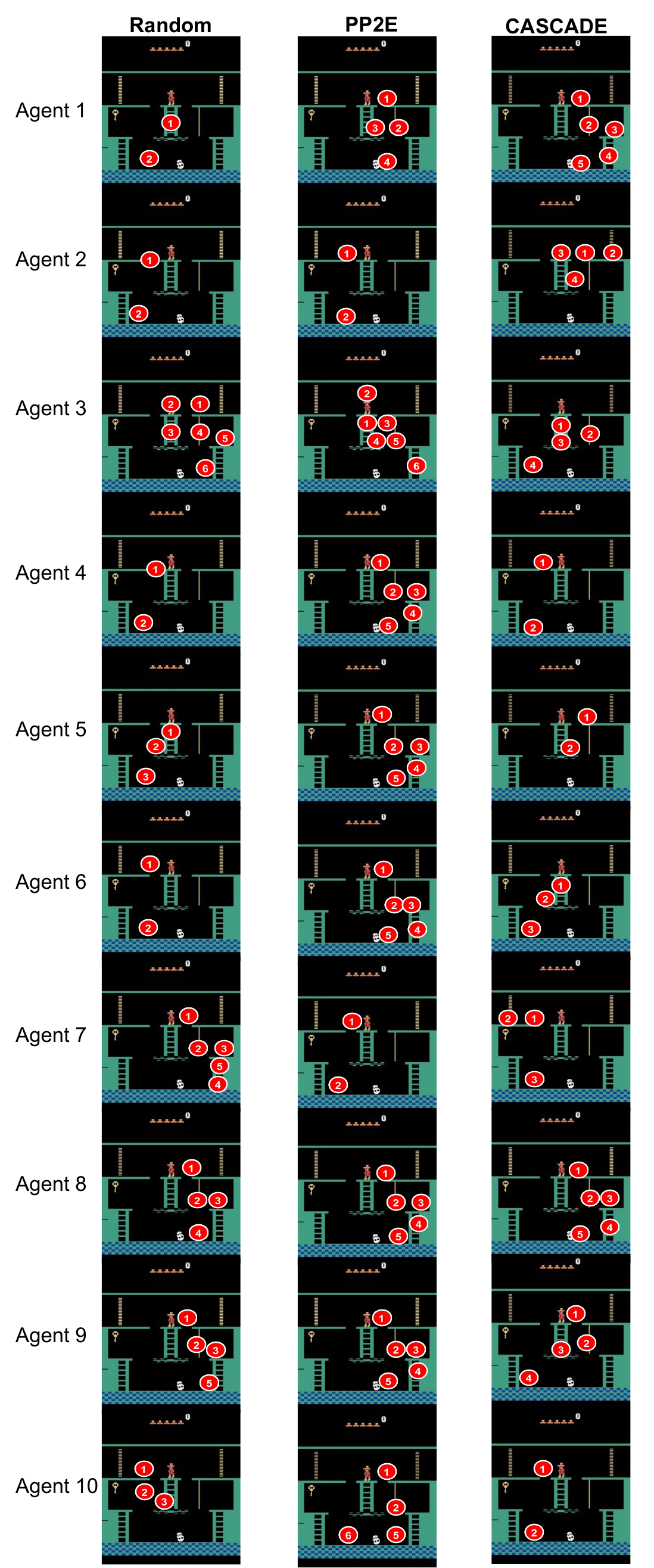}
    \caption{\small{\textbf{Trajectories in Montezuma's Revenge}: Each row shows a trajectory from one of the 10 exploration agents of a model after 3M training timesteps (P2E is omitted because it only has 1 agent.) We can see that PP2E's agents exhibit the most homogeneous behaviors, and result in trajectories that focus on the bottom right of the room, while CASCADE agents manage to explore more of the room collectively.}}
    \label{fig:atari_trajectory}
\end{figure}

To understand why \textsc{cascade} performs well, in Figure~\ref{fig:atari_trajectory} we plot trajectories from Montezuma's Revenge. We see that the inclusion of a diversity-inducing objective does indeed lead to more diverse behaviors in the environment, with each policy exploring different regions of the room. In contrast, for \textsc{pp2e}, agents 3,4,5,6,7,9 all exhibit very similar behavior, likely indicating that random initialization alone is not sufficient to induce diversity when all agents optimize for the same objective. 

\section{Implementation Details}
\label{app:imlementation_details}

DreamerV2 consists of an image encoder that uses a Convolutional Neural Network (CNN, \citep{cnn}), a Recurrent State-Space Model (RSSM \citep{planet}) that learns the dynamics, and predictors for the image, reward, and discount factor. The RSSM uses a sequence of deterministic recurrent states $h_t$. At each step, it computes a posterior state $z_t$ conditioned on the current image $x_t$, as well as a prior state $\hat{z_t}$ without the current image. During world model training, the concatenation of $h_t$ and $z_t$ is used to reconstruct the current image $x_t$, and predict the reward $r_t$ and discount factor $\gamma_t$. Once the world model is trained, it can be used to roll out ``imaginary" trajectories where the model state is instead the concatenation of the deterministic state $h_t$ and prior stochastic state $\hat{z_t}$.
\eq{
\text{RSSM} 
 \begin{cases}
\begin{alignedat}{3}
& \text{Recurrent model:}  \quad\quad\quad\quad && h_t &\ = &\ f_\psi(h_{t-1},z_{t-1},a_{t-1}) \\
& \text{Representation model:}    \quad\quad\quad\quad && z_t  &\ \sim &\ q_\psi(z_t | h_t,x_t) \\
& \text{Transition predictor:} \quad\quad\quad\quad && \hat{z}_t &\ \sim &\ p_\psi(\hat{z}_t | h_t) 
\end{alignedat}
\end{cases}
}

\eq{
\begin{alignedat}{4}
& \text{Image predictor:} \quad\quad\quad\quad\quad  && \hat{x}_t &\ \sim &\ p_\psi(\hat{x}_t | h_t,z_t) \\
& \text{Reward predictor:} \quad\quad\quad\quad\quad  && \hat{r}_t &\ \sim &\ p_\psi(\hat{r}_t | h_t,z_t) \\
& \text{Discount predictor:} \quad\quad\quad\quad\quad  && \hat{\gamma}_t &\ \sim &\ p_\psi(\hat{\gamma}_t | h_t,z_t).
\end{alignedat}
}

Our implementation was built on top of the official DreamerV2 repository (\url{https://github.com/danijar/dreamerv2}). We used the default hyperparameter values in the DreamerV2 repository. Table~\ref{table:hyperparameters} lists the additional hyperparameters used in our experiments. 

We also incorporate a latent disagreement ensemble, following Plan2Explore \citep{sekar2020planning}. This involves training, alongside the RSSM model, an MLP ensemble with 10 members (each having 4 hidden layers, 400 units per layer, and different parameter initializations). These one-step models take action $a_t$ and latents $z_t$, $h_t$ as inputs, and try to predict the next stochastic latent $z_{t+1}$. The variance across these outputs during imagined rollouts is then used as the reward signal that forms the Information Gain (\textcolor{blue}{$\mathrm{InfoGain}$}) objective. Note that this ensemble is otherwise unused, and is solely trained for the purposes of the exploration objective.

\begin{table}[H]
\caption{\small{\textsc{cascade} hyperparameters}}
\label{table:hyperparameters}
\begin{center}
\scalebox{0.85}{
\begin{tabular}{llc}
\toprule
\textbf{Environment} & \textbf{Parameter} & \textbf{Value} \\
\midrule
\multirow{5}{*}{MiniGrid-FourRooms} & \textsc{cascade} weight ($\lambda$) & 0.1 \\
                           & batch size & 200k \\
                           & deployments & 5 \\
                           & explorer train steps & 10k \\
                           & offline model train steps & 20k \\
\midrule
\multirow{5}{*}{MiniGrid-MultiRoom-N4S5} & \textsc{cascade} weight ($\lambda$) & 0.3 \\
                           & batch size & 200k \\
                           & deployments & 5 \\
                           & explorer train steps & 10k \\
                           & offline model train steps & 20k \\
\midrule
\multirow{5}{*}{Montezuma's Revenge} & \textsc{cascade} weight ($\lambda$) & 0.8 \\
                                     & batch size & 200k \\
                                     & deployments & 15 \\
                                     & explorer train steps & 10k \\
                                     & offline model train steps & 5k \\
\midrule
\multirow{6}{*}{Walker} & \multirow{2}{*}{\textsc{cascade} weight($\lambda$)} & 0.1 (Random) \\
                        &                                                   & 0.3 (Medium/Expert) \\
                        & batch size & 200k \\
                        & deployments & 2 \\
                        & explorer train steps & 250 \\
                        & offline model train steps & 5k \\
\bottomrule 
\end{tabular}
}
\end{center}
\end{table}

\newpage
\section{Deriving the Objective}
\label{app:deriving_obj}

\subsection{An Information Theoretic Perspective}\label{lemma::information_theoretic_perspective}

Inspired by~\cite{sekar2020planning}, we derive a data acquisition objective based on the mutual information between the collected data and parameters of the world model $\mathcal{M}_\psi$. $\mathcal{M}_\psi$ can be either stochastic or deterministic, and can also be an ensemble of empirical models. Crucially however, we eschew the reliance on one-step information gain (as performed in \cite{sekar2020planning}), and instead aim to maximize diversity directly in the space of trajectories. 
To do this, let $\Phi : \Gamma \rightarrow \Omega$ be an summary function mapping trajectories into an embedding `summary' space. For any model $\mathcal{M}_\psi$ we use the notation $\mathbb{P}_\pi[\mathcal{M}_\psi]$ to denote the distribution over trajectories generated by policy $\pi$ in $\mathcal{M}_\psi$ and use $\mathcal{M}_\psi \sim \mathbb{P}_{\mathcal{M}_\psi}$ to denote the posterior distribution over models. The later depends on the data collected so far. 

In this work we consider different embedding functions $\Phi$,

\begin{enumerate}
    \item \textbf{Final State Embedding}. In this setting we define $\Phi(\tau) = h_H$ where $h_H$ corresponds to the $H-$th (and last) state embedding in the trajectory $\tau$. Since we use an RNN world model, the use of this summary embedding is analogous to the final encoder embedding in $\mathrm{seq2seq}$ language models \cite{seq2seq}.
    \item \textbf{Visitation Frequencies}. In the tabular setting if we define $\Phi(\tau)$ as a discounted count statistic for states in $\tau$, we recover $\mathbb{P}_{\pi}^\Phi[\mathcal{M}_\psi] = d_{\mathcal{M}_\psi}^\pi$ where $d_{\mathcal{M}_\psi}^{\pi}$ corresponds to the discounted visitation distribution of policy $\pi$ in model $\mathcal{M}_\psi$.
\end{enumerate}

The authors of~\cite{sekar2020planning} study a per state-action mutual information objective that informs the construction of a greedy mutual information maximizing policy. Implicitly, this per-step objective assumes the model to factor in \emph{independent} transition operators pertaining to each state-action pair. This is certainly not the case when using powerful function approximators such as neural networks. In this case, the dynamics model corresponding to the state transitions of a specific state action pair is correlated with the state transition model of \emph{other} state-actions. This is not captured accurately by the objective in~\cite{sekar2020planning}. Instead, we consider the following choice for an exploratory policy in the single policy setting studied in~\cite{sekar2020planning}:

\begin{equation}\label{eq::single_policy_info_objective}
    \pi_{\mathrm{EXP}} = \max_{\pi} \mathcal{I}\left( \mathbb{P}_\pi^{\Phi}[\mathcal{M}_\psi] ; \mathcal{M}_\psi \right) = \mathcal{H}(\mathbb{P}_\pi^{\Phi}[\mathcal{M}_\psi]) - \mathcal{H}(\mathbb{P}_\pi^{\Phi}[\mathcal{M}_\psi] | \mathcal{M}_\psi).
\end{equation}

$\pi_{\mathrm{EXP}}$ is a policy whose distribution over summaries $\Phi(\tau)$ with $\tau \sim \mathbb{P}_\pi[\mathcal{M}_\psi]$ and $\mathcal{M}_\psi \sim \mathbb{P}_{\mathcal{M}_\psi}$ has large entropy, but such that the average entropy of the summaries in every individual model is small. The term $\mathcal{H}(\mathbb{P}_\pi^{\Phi}[\mathcal{M}_\psi])$ captures the total uncertainty (epistemic + aleatoric) while subtracting the conditional entropy removes the aleatoric uncertainty resulting from noise within the model and the policy.

To gain intuition about this objective, consider the case where $\Phi$ equals the \emph{final state embedding}. For simplicity, we consider the scenario when all models $\mathcal{M}_\psi \sim \mathbb{P}$ and all policies $\pi\sim \Pi$ are such that for any realization of $\mathcal{M}_\psi=m_\psi$, the conditional entropy $\mathcal{H}(\mathbb{P}_\pi^{\Phi}[\mathcal{M}_\psi] | \mathcal{M}_\psi = m_\psi) = \sigma(m_\psi)$ is a function of the world $m_\psi$ and not of the policy. In this case, the policy $\pi_{\mathrm{EXP}}$ is the entropy maximizing policy: 
    \begin{equation*}
    \pi_{\mathrm{EXP}} = \max_{\pi} \mathcal{H}(\mathbb{P}_\pi^{\Phi}[\mathcal{M}_\psi]).
\end{equation*}

The scenario above is realized for example when all models $\mathcal{M}_\psi \sim \mathbb{P}_{\mathcal{M}_\psi}$ are deterministic and all policies in $\Pi$ are deterministic. In this case, for every realization of $\mathcal{M}_\psi = m_\psi$, the conditional entropy $\mathcal{H}(\mathbb{P}_\pi^{\Phi}[\mathcal{M}_\psi] | \mathcal{M}_\psi = m_\psi) = 0$. The assumption $\mathcal{H}(\mathbb{P}_\pi^{\Phi}[\mathcal{M}_\psi] | \mathcal{M}_\psi = m_\psi) = \sigma(w)$ is also realized when the distribution $\Phi(\tau) \sim \mathbb{P}_{\pi}^\Phi[m_\psi]$ is approximated as a Gaussian distribution $\mathbb{P}_{\pi}^\Phi[m_\psi] \approx \mathcal{N}(  \mu(m_\psi, \pi  ) , \Sigma(m_\psi) ) $  with policy dependent mean $\mu(m_\psi, \pi)$ and policy independent covariance $\Sigma(m_\psi)$.

In these cases $\pi_\mathrm{EXP}$ is the policy that produces the most diverse distribution over final states across the posterior distribution over models $\mathbb{P}_{\mathcal{M}_\psi}$. Since access to $\mathcal{H}(\mathbb{P}_\pi^{\Phi}[\mathcal{M}_\psi])$ may not be possible, in our experiments we make further approximations inspired by~\cite{sekar2020planning}. Under the same gaussian approximation $\mathbb{P}_{\pi}^\Phi[m_\psi] \approx \mathcal{N}(  \mu(m_\psi, \pi  ) , \Sigma(m_\psi) ) $, optimizing $ \max_{\pi} \mathcal{H}(\mathbb{P}_\pi^{\Phi}[\mathcal{M}_\psi]) $ is achieved by finding the policy $\pi$ that makes the ensemble means $\mu(m_\psi, \pi)$ as far apart as possible. A suitable surrogate for this objective is to maximize the empirical variance over means, 

\begin{equation*}
    \mathrm{Var}(\pi) = \frac{1}{|\mathcal{M}_\psi|-1}\sum_{m_\psi} \| \mu(m_\psi, \pi) - \mu'(\pi)\|^2 
\end{equation*}

where $|\mathcal{M}_\psi|$ denotes the number of samples $m_\psi$, and $\mu'(\pi) = \frac{1}{|\mathcal{M}_\psi|} \sum_{m_\psi} \mu(m_\psi, \pi) $. We now consider a ``batch" version of Eq.~\ref{eq::single_policy_info_objective} involving a population of $B$ agents:

\begin{align}\label{equation::batch_mutual_info_objective_appendix}
     \{\pi^{(i)}_{\mathrm{EXP}}\}_{i=1}^B = \max_{\pi^{(1)}, \cdots, \pi^{(B)} \in \Pi^B} \mathcal{I}\left( \prod_{i=1}^B \mathbb{P}_{\pi^{(i)}}^{\Phi}[\mathcal{M}_\psi] ; \mathcal{M}_\psi \right) = \qquad \qquad \qquad \qquad \notag\\
     \mathcal{H}\left(\prod_{i=1}^B \mathbb{P}_{\pi^{(i)}}^{\Phi}[\mathcal{M}_\psi] \right) - \mathcal{H}\left(\prod_{i=1}^B \mathbb{P}_{\pi^{(i)}}^{\Phi}[\mathcal{M}_\psi]  \Big| \mathcal{M}_\psi \right)
\end{align}

where  $\prod_{i=1}^B \mathbb{P}_{\pi^{(i)}}^{\Phi}[\mathcal{M}_\psi] $ is the product measure of the candidate policies embedding distributions over the model $\mathcal{M}_\psi$. By definition the conditional entropy factors as,

\begin{equation}\label{equation::entropy_decomposition_product_appendix}
    \mathcal{H}\left(\prod_{i=1}^B \mathbb{P}_{\pi^{(i)}}^{\Phi}[\mathcal{M}_\psi] \Big| \mathcal{M}_\psi\right) = \sum_{i=1}^B \mathcal{H}\left(\mathbb{P}_{\pi^{(i)}}^{\Phi}[\mathcal{M}_\psi] \Big |  \mathcal{M}_\psi\right).
\end{equation}

The objective in Eq.~\ref{equation::batch_mutual_info_objective_appendix} has a similar interpretation as in the single policy case. We are hoping to find a set of policies whose average conditional entropies are small (Eq.~\ref{equation::entropy_decomposition_product}), but whose total entropies are large. Intuitively, we see this objective is more amenable to our population of policies. Concretely, by considering distributions directly over the space of trajectories, we ensure that each agent does not `double count' the explored states under Eq. \ref{equation::batch_mutual_info_objective_appendix}, whereas applying the same principal to a one-step information gain objective would simply ensure diversity \emph{conditioned} on that state and action, and doesn't explicitly ensure diversity in the visited states by the population.

\subsubsection{Proof of Lemma~\ref{lemma::all_policies_same_suboptimality}}
\label{sec:prooflemma1}

In the proof of Lemma~\ref{lemma::all_policies_same_suboptimality} we will make use of the following supporting result,

\begin{lemma}\label{lemma::entropy_supporting_lemma_one}
Let $p \in (0,1]$ and $p_1, p_2 > 0$ satisfying $p_1 + p_2 = p$ then,
\begin{equation*}
     p\log(1/p)< p_1 \log(1/p_1) + p_2 \log(1/p_2). 
\end{equation*}
\end{lemma}

\begin{proof}
Let $p_1 = \alpha p$ with $\alpha \neq 0,1$, then, 
\begin{align*}
   p_1 \log(1/p_1) + p_2 \log(1/p_2) &= \alpha p \log(1/(\alpha p)) + (1-\alpha) p \log(1/((1-\alpha)p)) \\
   &= p \log(1/p) + p \left(  \underbrace{ \alpha \log(1/\alpha) + (1-\alpha) \log(1/(1-\alpha)) }_{> 0} \right) \\
   &> p \log(1/p).
\end{align*}

\end{proof}

Lemma~\ref{lemma::entropy_supporting_lemma_one} implies the following result,

\begin{lemma}\label{lemma::entropy_supporting_lemma_two}
Let $\p \in [0,1]^K$ be a vector satisfying $\sum_{i=1}^K \p_i = 1$ and let $\mathcal{C}$ be a partition of $[K]$ such that $| \mathcal{C}  | \leq K-1 $. For all $C \in \mathcal{C}$ denote $\p(C) = \sum_{i \in C} \p_i$. The following inequality holds,
\begin{equation*}
    \sum_{C \in \mathcal{C}} \p(C) \log(1/\p(C)) \leq \sum_{i=1}^K \p_i \log(1/\p_i). 
\end{equation*}
If $\p_i > 0$ the inequality is strict,
\begin{equation*}
    \sum_{C \in \mathcal{C}} \p(C) \log(1/\p(C)) < \sum_{i=1}^K \p_i \log(1/\p_i). 
\end{equation*}

\end{lemma}

\begin{proof}
W.l.o.g we call $C_1$ the partition set containing $1$ and assume $\p_1 >0$. If $\p_i > 0$ for only one element of $C_1$, it must be the case that $\p_1 \log(1/\p_1)  = \p_{C_1}\log(1/\p_{C_1})$.

We now assume that $|C_1| > 1$ and that $C_1$ has at least two indices $i \neq j$ such that $\p_i , \p_j > 0$. Let $l$ denote the size of $C_1$ and w.l.o.g. let's say $C_1 = \{1, 2, \cdots, l\}$. Lemma~\ref{lemma::entropy_supporting_lemma_one} implies the following inequalities, 

\begin{align*}
    \p(C_1) \log(1/\p(C_1)) &< \p_1 \log(1/\p_1) + \p(C_1 \backslash \{1\}) \log(1/\p(C_1 \backslash \{1\}))\\
    &< \cdots \\
    &< \sum_{i \in C_1} \p_i \log(1/p_i). 
\end{align*}

Applying this reasoning to each element in the partition $\mathcal{C}$ yields the desired result.

\end{proof}

\lemmadiversityone*

\begin{proof}

First let's observe that when the models and policies are all deterministic, the conditional entropies satisfy,

\begin{equation*}
    \mathcal{H}\left(\prod_{i=1}^B \mathbb{P}_{\pi^{(i)}}^{\Phi}[\mathcal{M}_\psi] \Big| \mathcal{M}_\psi\right) = \sum_{i=1}^B \mathcal{H}\left(\mathbb{P}_{\pi^{(i)}}^{\Phi}[\mathcal{M}_\psi] \Big |  \mathcal{M}_\psi\right) = 0.
\end{equation*}

We assume the set of policies $\Pi$ is of size at least $B$. Otherwise, the result cannot be true. The first result follows immediately by Lemma~\ref{lemma::entropy_supporting_lemma_two}. For any fixed $\pi$, the product distribution $\prod_{i=1}^B \mathbb{P}_{\pi}^{\Phi}[{\mathcal{M}_\psi}]$ is a distribution over `diagonal' tuples of the form $\underbrace{(b, \cdots, b)}_{\text{size }B}$ where $b$ is an embedding. 

Consider an arbitrary set of policies $\pi^{(2)}, \cdots, \pi^{(B)}$ satisfying $\pi^{(i)} \neq \pi^{(j)} \neq \pi$ for all $i,j \in \{2, \cdots, K\}$. Call $\pi^{(1)} =  \pi$. The distribution over embeddings induced by the product distribution $\prod_{i=1}^B \mathbb{P}_{\pi^{(i)}}^{\Phi}[{\mathcal{M}_\psi}]$ is made of tuples of the form $(b_1,\cdots, b_B)$. Call $\p(b_1, \cdots, b_B)$ the probability under the product measure $\prod_{i=1}^B \mathbb{P}_{\pi^{(i)}}^{\Phi}[{\mathcal{M}_\psi}] $ of the tuple $(b_1,\cdots, b_B)$. Notice the `projection measure' onto the first coordinate satisfies 
\begin{equation*}
    \prod_{i=1}^B \mathbb{P}_{\pi^{(i)}}^{\Phi}[{\mathcal{M}_\psi}](b_1) = \prod_{i=1}^B \mathbb{P}_{\pi}^{\Phi}[{\mathcal{M}_\psi}](b_1, \cdots, b_1):= \p_1(b_1)
\end{equation*}
and that $\sum_{b_2, \cdots, b_K} \p(b_1, b_2, \cdots, b_K) = \p_1(b_1)$. This induces a partition over the outcome space $(b_1, \cdots, b_K)$ corresponding to $C(b_1) = \{ (b_1, b_2, \cdots, b_K)\}_{b_2, \cdots, b_K}$.

A direct use of Lemma~\ref{lemma::entropy_supporting_lemma_two} implies the entropy of the product distribution over finer grained tuples is larger than the entropy over the diagonal of the product distribution having all coordinates equal to each other. In fact this result also informs when the inequality will be strict. This happens when (for example the measure over $\mathcal{M}_\psi$ is the counting measure) there exists two worlds $\mathcal{M}_\psi^{(1)}$ and $\mathcal{M}_\psi^{(2)}$ such that the embedding tuples $(b^{(1)}_1, \cdots, b^{(1)}_B)$ and $(b^{(2)}_1,\cdots, b^{(2)}_B)$ satisfy $b_1^{(1)} = b_1^{(2)}$ and $(b_2^{(1)}, \cdots, b_B^{(1)}) \neq (b_2^{(2)}, \cdots, b_B^{(B)})$.

We will use this observation to prove the second claim. Consider a family of MDPs formed of depth $L$ binary trees. In this family, the paths leading to the $L-1$ layer are the same, but the connections from layer $L-1$ to layer $L$ are unknown. The `posterior' distribution is assumed to be uniform over all plausible trees. Layer $L-1$ has size $2^{L-1}$ and layer $L$ has size $2^L$. W.l.o.g. we assume the set of nodes in layer $L$ is labeled as $[1,2,\cdots , 2^L]$. The distribution over models is supported over the set of partitions of size $2^{L-1}$ of $[1, \cdots, 2^{L}]$ where each partition set has size $2$. We assume $2^{L} \geq B$ so that there are at least $B$ distinct policies. 

Let $\pi$ be a fixed policy. It is enough to show there exist two worlds $\mathcal{M}_{\psi}^{(1)}$ and $\mathcal{M}_\psi^{(2)}$ such that policy $\pi$ ends in the same state for both $\mathcal{M}_{\psi}^{(1)}$ and $\mathcal{M}_\psi^{(2)}$ but that there exist distinct policies $\pi^{(2)}, \cdots, \pi^{(B)}$ such that their endpoints $(b_2^{(1)}, \cdots, b_B^{(1)})$ and $(b_2^{(2)}, \cdots, b_B^{(2)})$ in worlds $\mathcal{M}_{\psi}^{(1)}$ and $\mathcal{M}_\psi^{(2)}$ satisfy $(b_2^{(1)}, \cdots, b_B^{(1)}) \neq (b_2^{(2)}, \cdots, b_B^{(2)})$. The latter always holds because among the set of models that maintain the same endpoint for $\pi$, there is a pair of models that has different endpoints for $\pi^{(2)}$ for any arbitrary $\pi^{(2)} \neq \pi$. This suffices to show the entropy of the ensemble of distinct policies is strictly larger than the entropy of the `diagonal choice' $\underbrace{(\pi, \cdots, \pi)}_{\text{size }B}$.

\end{proof}

\subsubsection{InfoCascade}

Since the mutual information objective in Eq.~\ref{equation::batch_mutual_info_objective} is submodular, a simple greedy algorithm yields a $(1-\frac{1}{e})$ approximation of the optimum \cite{nemhauser1978analysis}. This is the same observation that gives rise to the greedy algorithm underlying other batch exploration objectives, such as in BatchBALD~\cite{kirsch2019batchbald}. 

Let's start by assuming we have candidate policies $\pi^{(1)}, \cdots, \pi^{(i-1)}$. InfoCascade selects $\pi^{(i)}$ based on the following greedy objective,

\resizebox{.99\linewidth}{!}{
  \begin{minipage}{\linewidth}
    \begin{align*}
        \pi^{(i)} &= \arg\max_{\tilde \pi^{(i)} \in \Pi} \mathcal{I}\left( \prod_{j=1}^i \mathbb{P}_{\tilde \pi^{(j)}}^{\Phi}[{\mathcal{M}_\psi}] ; {\mathcal{M}_\psi} \Big|\tilde \pi^{(j)} = \pi^{(j)}~~ \forall j \leq i-1 \right) \label{equation::greedy_information_theoretic_objective_2}\\
        &=\mathcal{H}\left(\prod_{j=1}^i \mathbb{P}_{\tilde \pi^{(j)}}^{\Phi}[{\mathcal{M}_\psi}] \Big| \tilde\pi^{(j)} = \pi^{(j)}~~ \forall j \leq i-1 \right) - \mathcal{H}\left(\prod_{j=1}^i \mathbb{P}_{\pi^{(j)}}^{\Phi}[\mathcal{M}_\psi]  \Big| \mathcal{M}_\psi , \tilde\pi^{(j)} = \pi^{(j)}~~ \forall j \leq i-1\right).
    \end{align*}
  \end{minipage}
}

Equation~\ref{equation::entropy_decomposition_product} implies \begin{align}
\mathcal{H}\left(\prod_{j=1}^i \mathbb{P}_{\pi^{(j)}}^{\Phi}[{\mathcal{M}_\psi}]  \Big| {\mathcal{M}_\psi} , , \tilde\pi^{(j)} = \pi^{(j)}~~ \forall j \leq i-1\right) = \sum_{j=1}^i \mathcal{H}\left(\mathbb{P}_{\pi^{(i)}}^{\Phi}[{\mathcal{M}_\psi}] \Big |  {\mathcal{M}_\psi}\right)
\end{align}
and therefore if we approximate $\mathbb{P}_{\pi}^\Phi[m_\psi] $ by a Gaussian $\mathbb{P}_{\pi}^\Phi[m_\psi] \approx \mathcal{N}(  \mu(m_\psi, \pi  ) , \Sigma(m_\psi) ) $, the conditional entropy becomes a policy independent term. In this case finding $\pi^{(i)}$ boils down to solving for the policy that maximizes $\mathcal{H}\left(\prod_{j=1}^i \mathbb{P}_{\pi^{(j)}}^{\Phi}[\mathcal{M}_\psi] \Big| \pi^{(1)}, \cdots, \pi^{(i-1)} \right)$. Using the same approximations described for the single policy objective, a suitable surrogate for this objective is to maximize $\max_\pi \mathrm{Var}(\pi | \pi^{(1)}, \cdots, \pi^{(i-1)})$, the empirical variance over means with respect to the policies found so far $\{ \pi^{(j)}\}_{j=1}^{i-1}$,

\begin{align*}
     \mathrm{Var}(\pi | \pi^{(1)}, \cdots, \pi^{(i-1)}) =\qquad \qquad \qquad \qquad \qquad \qquad \qquad \qquad \qquad \qquad \qquad \qquad \qquad \qquad \\
     \frac{1}{|\mathcal{M}_\psi||i|-1} \sum_{m_\psi} \sum_{\tilde \pi \in \{ \pi^{(1)}, \cdots, \pi^{(i-1)}, \pi\}} \| \mu(m_\psi, \tilde \pi) - \mu'(\pi, \pi^{(1)}, \cdots, \pi^{(i-1)}) \|^2   
\end{align*}

Where $\mu'( \pi, \pi^{(1)}, \cdots, \pi^{(i-1)}) = \frac{1}{|\mathcal{M}_\psi  | |i|} \sum_{m_\psi} \sum_{\tilde \pi \in \{ \pi^{(1)}, \cdots, \pi^{(i-1)}, \pi\}} \mu(m_\psi, \tilde \pi)  $.

\newpage
\section{Theoretical RL Intuition}
\label{app:theo_rl}

The problem we study in this work can be thought of as the ``batch" version of the Reward Free Exploration formalism~\cite{jin2020reward}. In this setting, the learner interacts with an MDP in two phases: 1) a training phase, where it is allowed to collect data from the environment; 2) a test phase, where the learner is presented with an arbitrary task (parametrized by a reward function unseen during training) and it must produce a near optimal policy. When faced with this problem, the learner is required to design a careful exploration strategy that permits them to build an accurate model in all areas of the state-space that can be reached with sufficient probability. There have been multiple follow up works that have also studied this problem in the context of Linear MDPs~\cite{wang2020reward} and neural function approximation~\cite{qiu2021reward}. Nonetheless, there has been limited focus on the batch setting, where the learner is required to collect data via parallel data gathering policies in each training phase.   

Due to the adaptive nature of the algorithm (every single environment query uses all information collected so far), there is an inevitable drop in performance when moving from the sequential ( fully online) setting to the batch (deployment efficient) setting, as measured by the total number of environment interactions (in our case, $T \times B$ where $T$ is the number of training rounds and $B$ the number of parallel collection policies). To illustrate how \textsc{cascade} mitigates this loss in sample-efficiency, we outline the pseudo-code of \textsc{cascade-ts}, a greedy Thompson Sampling algorithm (see \cite{agrawal2012analysis}) that produces the $i-$th batch exploration policy in a tabular enviornment by incorporating fake count data from rolling out policies $\pi^{(1)}, \cdots, \pi^{(i-1)}$ in the model.  In the algorithms we present in the main body, we incorporate data gathered in the model by previous $i-1$ selected policies into the batch when building the $i-$th exploration policy; this is directly related to the approach behind \textsc{cascade-ts}. Concretely, encouraging policy $\pi^{(i)}$ to induce a high variance among the embeddings produced by $\{\pi^{(1)}, \cdots, \pi^{(i-1)}\}$ can be thought as adding a bonus to encourage $\pi^{(i)}$ to visit regions of the space that have a low embedding visitation count by the previous policies in $\{\pi^{(1)}, \cdots, \pi^{(i-1)}\}$. 

\begin{algorithm}[!h]
  \caption{\textsc{cascade-ts}}\label{alg:diversity_sequential_algorithm}
    \begin{algorithmic}[1] 
        \STATE \textbf{Input:} Exploration batch size $B$. Fake samples parameter $M$.
        \STATE Initialize model $\hat{\mathbb{P}}_0$ over $\mathcal{S} \times \mathcal{A}$ state actions.
        \STATE Initialize the batch collection policies as $\{ \pi_0^{(i)}\}_{i=1}^B$ to uniform.
      \FOR{time in $k=0, 1, 2, \dots $}
          \STATE Collect trajectory data $\{\tau_k^{(i)}\}_{i=1}^B$ from $\pi^{(i)}_k$ for all $i =1, \cdots, B$. Update $\mathcal{D}^{k+1} \leftarrow \mathcal{D}^{k} \cup \{ \tau_k^{(i)}\}_{i=1}^B$.
          \STATE Initialize the fake data buffer $\mathcal{D}_{\mathrm{fake}}^{k+1}= \emptyset$.
          \FOR{$i=1, \cdots, B$}
              \STATE Sample MDP model from posterior $\widetilde{\mathbb{P}}_{k+1}^{(i)} \sim \mathbb{P}(\cdot | \mathcal{D}^{k+1} \cup \mathcal{D}_{\mathrm{fake}}^{k+1} )$.
              \STATE Compute fake counts $N^{(i)}_{k+1}(s,a) = \sum_{(\tilde{s}, \tilde{a},\tilde{s}') \in \mathcal{D}^{k+1} \cup \mathcal{D}_{\mathrm{fake}}^{k+1}}  \mathbf{1}(\tilde{s} = s, \tilde{a} = a)  $.
              \STATE Compute fake uncertainty bonuses $b_{k+1}^{(i)}(s,a) \propto \sqrt{\frac{1}{N^{(i)}_{k+1}(s,a)}}$
              \STATE Solve for $\pi_{k+1}^{(i)}$ by solving,
              \begin{equation*}
                  \pi_{k+1}^{(i)} = \max_{\pi } \mathbb{E}_{\tau \sim \widetilde{\mathbb{P}}_{k+1}^{(i), \pi} }\left[ \sum_{h=1}^H  b_{k+1}^{(i)}(s_h, a_h)  \right] 
              \end{equation*}
                  
              \STATE Collect fake trajectory data $\{ \tau^{(i)}_{\ell, \mathrm{fake}} \}_{\ell=1}^{M}$ from posterior sampled model $\widetilde{\mathbb{P}}_{k+1}^{(i)}$.
              \STATE Update fake data buffer $\mathcal{D}_{\mathrm{fake}}^{k+1} \leftarrow \mathcal{D}_{\mathrm{fake}}^{k+1} \cup \{ \tau_{\ell,  \mathrm{fake}}^{(i)}\}_{\ell=1}^M$.
          \ENDFOR
       \ENDFOR
       
  \end{algorithmic}
\end{algorithm}

Algorithm~\ref{alg:diversity_sequential_algorithm} works by sampling a model from a model posterior, solving for an optimal uncertainty seeking policy and updating the model with `fake' data corresponding to trajectories collected in the model from this uncertainty seeking policy. After producing the $i-$th exploration policy in the batch, the recomputed uncertainty bonuses of the model are reduced in the areas of the state space most visited by the $i-$th policy. This will encourage subsequent uncertainty seeking policies (i.e., $\pi^{(i+1)}_{k+1}, \pi^{(i+2)}_{k+1},\dots,\pi^{(B)}_{k+1}$) to visit parts of the space not yet explored by previous policies. 

\textbf{Why Thompson Sampling?} The reader may wonder why Algorithm~\ref{alg:diversity_sequential_algorithm} samples a model $\widetilde{\mathbb{P}}_{k+1}^{(i)} $ from a TS posterior instead of using the empirical model resulting from fake and true data. The answer is that the raw empirical model may ascribe a probability of zero to certain transitions, which means we may miss exploring parts of the true state-action space. Having a TS prior that ascribes non-zero probabilities to all possible transitions fixes this potential issue.

\subsection{Proof of Lemma~\ref{lemma::cascade_TS}}

Let's consider the class of deterministic MDPs with $S = |\mathcal{S}|$ states and $A = |\mathcal{A}|$ actions. It is clear that playing the same deterministic policy multiple times does not provide us with any more information than playing it once. We will then show that running Thompson TS, the learner will end up with a nonzero probability of producing at least two distinct policies within a batch of size $B > 2$. 

We will further assume the posterior is aware of the deterministic nature of the MDP family so that any sample MDP model $\widetilde{\mathbb{P}}_{k+1}^{(i)}$ is deterministic. For $i > 1$, the model $\widetilde{\mathbb{P}}_{k+1}^{(i)}$ is one whose transitions are consistent with the fake data generated by $\pi_{k+1}^{(1)}, \cdots, \pi_{k+1}^{(i-1)}$ (and the true data in $\mathcal{D}^{k}$).

A deterministic MDPs can be encoded as a set of triplets $(s, a, s')$. We say a model $\{ (s,a,s') \}_{s \in \mathcal{S}, a \in \mathcal{A}}$ is $\epsilon-$accurate if at least a fraction of $1-\epsilon$ of the triplets are correct. 

We now illustrate how the \textsc{cascade-ts} algorithm evolves in this setting. To simplify our task, we will further restrict our attention to the family of deterministic MDPs made of binary trees of height $L$. The number of nodes in any of such trees equals $2^{L+1}-1$.We will define the counts to be a large value ($2L$) when no data has been collected of a given state action pair. When data has been collected, we define the bonus terms to be of order at most $1$.

To figure out the connectivity of any such trees it is necessary to try out $2^L-1$ distinct sequences of $L$ actions (the nature and connectivity of the remaining leaf nodes can be inferred once all the other ones have been decoded).  To build an $\epsilon-$accurate model, it is enough to know the connectivity structure of $1-\epsilon$ proportion of paths corresponding to $(1-\epsilon)(2^L-1)$ leaf nodes.

To prove the result of Lemma~\ref{lemma::cascade_TS}, we observe that any batch strategy can be simulated by a fully sequential learner, so it immediately follows that $  T(\epsilon, \mathrm{Sequential}) \leq   T(\epsilon, \mathrm{AnyBatchStrategy}) $. To show that \textsc{cascade-ts} will be better than $\mathrm{SinglePolicyBatch}$ in this tree example, as long as $B$ is smaller than the remaining number of paths the learner has not tried out, it will produce a set of $B$ policies different from any policy played so far and also different from each other. 

To see why the second inequality is true, it is enough to see that complete paths (real or imagined) have a total reward score of at most $L$, since the counts are always at most one for each edge that is present. Nonetheless, the counts over any path (sequence of $L$ actions) that has not been visited nor has been imagined to be visited will be at least $2L$. Thus, every step in the sequential batch construction mechanism will produce a new policy (sequence of $L$ actions) not existent neither in the true nor the fake data so far. This finalizes the proof. Notice that the $\mathrm{SinglePolicyBatch} $ strategy will have tried only $K$ unique paths after $K$ batches have been collected, whereas the \textsc{cascade-ts} strategy will have tried $B K$, resulting in a potentially substantial improvement in sample efficiency.

\end{document}